\documentclass[lettersize,journal]{IEEEtran}
\usepackage{amsmath,amsfonts}
\usepackage{amsthm}
\usepackage{amssymb}
\usepackage{algorithm}
\usepackage{algpseudocode}
\usepackage{array}
\usepackage[caption=false,font=normalsize,labelfont=sf,textfont=sf]{subfig}
\usepackage{textcomp}
\usepackage{stfloats}
\usepackage{url}
\usepackage{verbatim}
\usepackage{graphicx}
\usepackage{cite}
\usepackage{epstopdf}
\usepackage{enumitem}

\usepackage{tcolorbox}

\DeclareMathOperator*{\argmin}{argmin}

\begin{document}

\title{GenSafe: A Generalizable Safety Enhancer for Safe Reinforcement Learning Algorithms Based on Reduced Order Markov Decision Process Model}

\author{Zhehua Zhou, Xuan Xie, Jiayang Song, Zhan Shu, and Lei Ma
\thanks{Zhehua Zhou, Xuan Xie, Jiayang Song, Zhan Shu are with University of Alberta, Canada (email: zhehua1@ualberta.ca, xxie9@ualberta.ca, jiayan13@ualberta.ca, zshu1@ualberta.ca).}
\thanks{Lei Ma is with The University of Tokyo, Japan and University of Alberta, Canada (email: ma.lei@acm.org).}
\thanks{This paper has been accepted by IEEE Transactions on Neural Networks and Learning Systems in Oct, 2024.}
}

\markboth{IEEE Transactions on Neural Networks and Learning Systems}%
{Zhou \MakeLowercase{\textit{et al.}}: GenSafe: A Generalizable Safety Enhancer for Safe Reinforcement Learning Algorithms Based on Reduced Order Markov Decision Process Model}


\maketitle

\begin{abstract}
Safe Reinforcement Learning (SRL) aims to realize a safe learning process for Deep Reinforcement Learning (DRL) algorithms by incorporating safety constraints. 
However, the efficacy of SRL approaches often relies on accurate function approximations, which are notably challenging to achieve in the early learning stages due to data insufficiency.
To address this issue, we introduce in this work a novel Generalizable Safety enhancer (GenSafe) that is able to overcome the challenge of data insufficiency and enhance the performance of SRL approaches. 
Leveraging model order reduction techniques, we first propose an innovative method to construct a Reduced Order Markov Decision Process (ROMDP) as a low-dimensional approximator of the original safety constraints. 
Then, by solving the reformulated ROMDP-based constraints, GenSafe refines the actions of the agent to increase the possibility of constraint satisfaction.
Essentially, GenSafe acts as an additional safety layer for SRL algorithms. 
We evaluate GenSafe on multiple SRL approaches and benchmark problems. 
The results demonstrate its capability to improve safety performance, especially in the early learning phases, while maintaining satisfactory task performance.
Our proposed GenSafe not only offers a novel measure to augment existing SRL methods but also shows broad compatibility with various SRL algorithms, making it applicable to a wide range of systems and SRL problems.
\end{abstract}


\section{Introduction}
\label{sec.introduction}
\IEEEPARstart{O}{ver} the past decade, Deep Reinforcement Learning (DRL) has achieved remarkable advancements in the control of a diverse array of autonomous systems, such as robotic manipulators~\cite{liu2021deep}, autonomous vehicles~\cite{kiran2021deep}, and drones~\cite{azar2021drone}. 
However, the inherent random exploration characteristic of DRL algorithms often undermines the safety of the learning process~\cite{garcia2015comprehensive}. 
During the progression towards an optimal policy, the agent may exhibit numerous unsafe intermediate policies, posing risks not only to the system itself but also to its surrounding environment~\cite{ladosz2022exploration}.
For example, an autonomous vehicle might repeatedly collide with obstacles prior to the successful determination of a DRL-based controller. 
Due to these safety concerns, the majority of DRL research findings have been confined to simulated environments~\cite{duan2016benchmarking}, with fewer applications in real-world experiments~\cite{zhou2020general}. 
This limitation becomes more evident in tasks where ensuring safety is of paramount importance.

To address this challenge, a variety of Safe Reinforcement Learning (SRL) approaches have been proposed~\cite{brunke2022safe,gu2022review}, aiming to realize a safe learning process for DRL algorithms. 
Since incorporating safety specifications naturally involves constraints, state-of-the-art SRL approaches often employ the Constrained Markov Decision Process (CMDP)~\cite{altman2021constrained} as their fundamental mathematical framework.
In CMDP, a cost function is defined in addition to the standard reward function, which maps undesirable behavior at any given time step to a non-negative scalar cost.
A system is considered safe if a cost-related constraint, e.g., the accumulated cost should remain below a predefined threshold, is satisfied~\cite{wachi2020safe}. 
This transforms the SRL problem into a constraint satisfaction problem and allows for the independent consideration of task performance and safety specifications~\cite{ray2019benchmarking}.
However, although optimal solutions for finite and low-dimensional CMDPs can be obtained by linear programming~\cite{vanderbei2020linear}, solving CMDPs for complex systems with high-dimensional and continuous state and action spaces remains a challenging problem.

Modern SRL techniques typically fall into two categories: model-based and model-free approaches~\cite{gu2022review}. 
Model-based approaches primarily focus on learning a system or cost model, which is then utilized to refine the current policy~\cite{liu2020safe,cowen2022samba,thomas2021safe,jayant2022model,sikchi2022learning,ma2022conservative,as2022constrained,wen2018constrained}. 
Although these approaches generally exhibit better sample efficiency compared to model-free ones, their performance is highly dependent on the accuracy of the learned model~\cite{jayant2022model}.
Conversely, model-free approaches usually employ policy gradient methods, necessitating function approximations for both reward and cost functions~\cite{achiam2017constrained,tessler2018reward,yu2019convergent,zhang2020first}.
Accurate approximations are crucial in these methods for ensuring both monotonic performance improvement and constraint satisfaction~\cite{achiam2017constrained}.
However, obtaining reliable and precise function approximations is a challenging task, especially in the early learning stage, where the amount of available data is often insufficient.
This inevitably diminishes the efficacy and safety performance of model-free SRL methods~\cite{zhang2020first}.

\begin{figure}
    \centering
    \includegraphics[trim={1mm 8mm 1mm 2mm}, width=.9\linewidth]{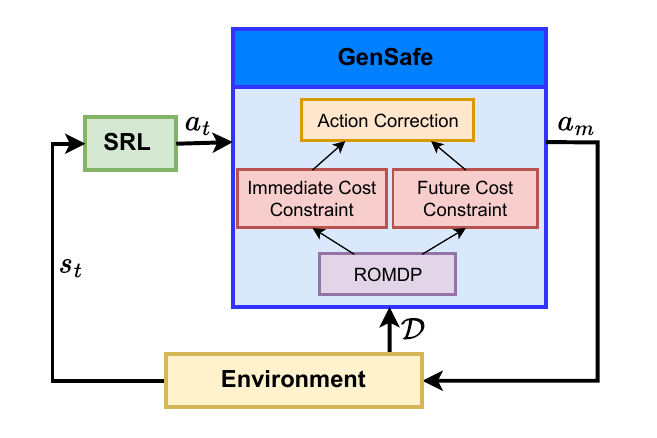}
    \caption{SRL with GenSafe. At each timestep, the current SRL policy recommends an action $a_t$ based on the current state $s_t$. Then, the proposed GenSafe performs an action correction to identify a modified action $a_m$ that is more likely to satisfy safety constraints. Such a corrective process involves resolving an optimization problem that considers both the immediate and future cost constraints, which are derived from the constructed ROMDP. We utilize the set of data samples $\mathcal{D}$ observed during the learning process to construct the ROMDP, which serves as a low-dimensional approximator of the original cost function in CMDP.}
    \label{fig.overview}
    \vspace{-10pt}
\end{figure}

To address this problem, we propose in this work a novel \textbf{Gen}eralizable \textbf{Safe}ty enhancer (GenSafe) that is able to improve the safety performance of model-free SRL algorithms, especially in early learning stages when data availability is limited.
The core idea of GenSafe is to mitigate data insufficiency by employing model order reduction techniques~\cite{schilders2008model,zhou2021learning,chen2020optimal} to derive a representative low-dimensional safety predictor model, which is then used to estimate the safety of the original system. 
This is achieved by first converting the original high-dimensional state space into a more manageable low-dimensional state space through dimensionality reduction. 
Subsequently, leveraging this reduced state space, we develop a novel approach to systematically construct a Reduced Order Markov Decision Process (ROMDP) model that serves as a low-dimensional approximator of the original cost function in CMDP.
By assuming that high-dimensional state and action pairs mapping to the same low-dimensional pairs also exhibit similar safety characteristics, the ROMDP is capable of efficiently predicting actual cost returns and system safety for unobserved control commands based on limited available data.
Such predictions offer guidance for optimizing agent actions to minimize the risk of constraint violations.
Utilizing these ROMDP-derived safety estimates, GenSafe incorporates an action correction mechanism into existing SRL algorithms. 
By resolving safety constraints that are redefined via the ROMDP, GenSafe modifies the actions proposed by the SRL agent at each learning timestep to increase the probability of constraint satisfaction, thereby improving the safety performance.
In essence, GenSafe acts as an additional safety layer for SRL algorithms (see Fig.~\ref{fig.overview}).

It is important to note that GenSafe does not represent a new SRL algorithm.
Instead, it is an innovative measure designed to augment the safety performance of existing SRL algorithms, particularly during their early learning phases.
Existing research on enhancing safety of the learning process typically relies on revising action commands based on either offline-collected safety data~\cite{dalal2018safe} or a comprehensive system model~\cite{fisac2018general,chow2018lyapunov,zanon2020safe}. 
Compared to these methods, GenSafe operates as a purely online method, obviating the requirement for any prior knowledge of the system model or SRL task. 
The construction of the ROMDP and the correction of actions are exclusively based on data collected throughout the learning process. 
Moreover, utilizing the proposed order reduction technique, GenSafe is able to provide valuable safety estimates and rectify control commands to enhance system safety, even in scenarios of data insufficiency. 
Additionally, designed as an extra safety layer, GenSafe demonstrates high generalizability and compatibility with a broad spectrum of SRL algorithms and tasks, highlighting its potential as a versatile tool for enhancing safety across diverse SRL applications. 
Detailed contributions of this work are summarized as follows.

\begin{itemize}[leftmargin=*]
    \item Based on order reduction techniques, we propose an innovative method to systematically construct a ROMDP model that serves as a low-dimensional approximator of the cost function in CMDP. 
    Such a ROMDP is able to effectively provide accurate estimates of system safety even in early learning stages when data is insufficient, thereby offering crucial guidance for improving constraint satisfaction.

    \item Utilizing the constructed ROMDP, we present GenSafe, a novel approach for augmenting existing SRL algorithms. 
    By modifying the actions of the agent, GenSafe enhances safety performance while maintaining task performance at a satisfactory level.
    Moreover, GenSafe is designed as an additional safety layer, which enables its compatibility and generalizability across various SRL algorithms and tasks.
    
    \item We conduct a comprehensive analysis of GenSafe's performance through experiments on a variety of SRL algorithms and benchmark problems.
    The results validate the effectiveness of GenSafe, supporting its use as a universal strategy for enhancing the safety performance of diverse SRL algorithms.
    
\end{itemize}

\section{Related Work}
\label{sec.related_work}

\subsection{Model-free SRL}

In the literature, numerous approaches have been proposed to address CMDPs~\cite{abad2002self,borkar2005actor,bhatnagar2012online} and SRL problems~\cite{hao2024exploration,ijcai2023p763,kim2024nips,zhang2022barrier}. 
Among these, Constrained Policy Optimization (CPO)~\cite{achiam2017constrained} is recognized as one of the most influential model-free approaches that employ the policy gradient method.
It utilizes surrogate functions to approximate both objective and constraint functions. 
However, the inherent difficulty in achieving accurate function approximation diminishes the performance of CPO in certain state-of-the-art SRL benchmark problems, such as SafetyGym~\cite{ray2019benchmarking}.
In~\cite{ray2019benchmarking}, a Lagrangian relaxation of the SRL problem is introduced, which is integrated with Proximal Policy Optimization (PPO)~\cite{schulman2017proximal} to create a PPO-Lagrangian algorithm, and with Trust Region Policy Optimization (TRPO)~\cite{schulman2015trust} for a TRPO-Lagrangian algorithm. 
Such a primal-dual method transforms the constrained problem into an unconstrained one by augmenting the objective with a sum of constraints, each weighted by its respective Lagrange multiplier~\cite{bertsekas2014constrained}.
Based on this concept, various primal-dual SRL approaches have been developed~\cite{tessler2018reward, yu2019convergent,yang2020projection,zhang2020first,wcsac,cup}.
For instance, derived from CPO, \cite{yang2020projection} introduces the Projection-based Constrained Policy Optimization (PCPO), a two-step method that first solves the policy search problem via TRPO and then projects the policy back to a feasible region for satisfying safety constraints.
Following this, \cite{zhang2020first} presents the First Order Constrained Policy Optimization (FOCOPS), which utilizes first-order optimization to achieve improved computational efficiency over PCPO.
\cite{wcsac} introduces an algorithm called Worst-Case Soft Actor Critic (WCSAC), which extends the soft actor critic algorithm with a safety critic to achieve risk control.
In~\cite{cup}, a policy optimization method based on Constrained Update Projection (CUP) is proposed, which employs surrogate functions derived from Generalized Advantage Estimation (GAE)~\cite{schulman2015high}.
However, these primal-dual methods often exhibit sensitivity to the initialization of Lagrange multipliers and learning rates, which could result in considerable hyperparameter tuning efforts~\cite{xu2021crpo}. 
Moreover, the introduction of dual parameters may pose stability challenges at the algorithm's saddle points~\cite{gu2022review}.

Another category of model-free SRL approaches is the primal methods~\cite{dalal2018safe,chow2018lyapunov,liu2020ipo}, which achieve constraint satisfaction through diverse designs of the objective function or the update process without the introduction of dual variables.
For example, \cite{liu2020ipo} introduces Interior Point Optimization (IPO), which utilizes logarithmic barrier functions to satisfy safety constraints. 
Similarly, Lyapunov functions have been employed in several SRL studies to maintain safety during the learning process~\cite{chow2018lyapunov}, where the agent's actions are restricted by these functions.
Nevertheless, these primal methods often require prior system knowledge for designing the barrier or Lyapunov functions, which can be inherently complex. 
Akin to our approach, \cite{dalal2018safe} also adds a safety layer to achieve safe exploration during training. 
It constructs a linear cost model and then solves a quadratic programming problem to determine a safe action. 
However, \cite{dalal2018safe} relies on pre-collected offline data to build the linear cost model, requiring new datasets for each task, which limits its applicability and generalizability. 
In contrast, our proposed GenSafe operates as an online learning method and does not require prior knowledge about the system or specific tasks.

A notable challenge with model-free SRL approaches is their dependence on accurate function approximations, e.g., for cost and reward functions, to achieve satisfactory performance~\cite{gu2022review,brunke2022safe}. 
This issue becomes more pronounced in the early learning phase, where limited data availability often hinders the creation of precise approximations, leading to reduced performance. 
To tackle this problem, we present GenSafe in this work, which utilizes a ROMDP to estimate the cost return more effectively. 
By employing order reduction, it mitigates the problem of data insufficiency, and as a result, improved constraint satisfaction is realized, particularly in the early learning stages.

\subsection{Model-based SRL}

Model-based SRL approaches~\cite{liu2020safe,cowen2022samba,jayant2022model,sikchi2022learning,ma2022conservative,as2022constrained,wen2018constrained} are generally considered more sample efficient than model-free ones.
In~\cite{liu2020safe}, a model-based method that learns both the system dynamics and the cost function for addressing safety concerns is proposed.
This method leverages the learned model to generate sample trajectories, which are then used to refine the policy via an adapted cross-entropy method. 
Similarly, \cite{cowen2022samba} employs PILCO~\cite{deisenroth2011pilco} for learning model dynamics and optimizes the policy using constraint functions based on Conditional Value at Risk (CVaR)~\cite{chow2014algorithms}.
However, the effectiveness of model-based SRL approaches often highly depends on the accuracy of the learned model. 
In practical tasks, obtaining an accurate system model could be a challenging task.

There exist also model-based SRL approaches~\cite{fisac2018general,berkenkamp2015safe,berkenkamp2017safe,zanon2020safe,lutjens2019safe} that are grounded in control-theoretical concepts or learning-based control~\cite{yang2022hamiltonian,yang2023cooperative,yang2021model}.
For example, in~\cite{fisac2018general}, reachability analysis is employed to devise a reliable backup policy that could rectify potentially hazardous actions taken by the agent.
Lyapunov functions are also frequently used to maintain system stability during the learning process~\cite{berkenkamp2015safe}. 
Nevertheless, these approaches often necessitate the manual design of Lyapunov functions, which might be infeasible for complicated tasks.
In \cite{berkenkamp2017safe}, the concept of region of attraction is utilized to define a safe region, and the exploration during the learning process is confined to this specified region.
Model Predictive Control (MPC)~\cite{kouvaritakis2016model} is another tool widely used for safe learning~\cite{zanon2020safe,lutjens2019safe}. 
These methods involve learning a system or disturbance model and applying robust MPC to determine safe actions~\cite{aswani2013provably}. 
However, approaches guided by control-theoretical concepts usually face computational difficulties~\cite{fisac2019bridging}, limiting their applicability predominantly to systems characterized by linear or low-dimensional dynamics.

\section{Preliminary}
\label{sec.preliminary}

In this section, we first present fundamental concepts of CMDP (Sec.~\ref{subsec.prel.cmdp}).
Then, we define the SRL problem within the CMDP framework (Sec.~\ref{subsec.prel.srl}).
Finally, we introduce PPO-Lagrangian, which is usually regarded as a baseline method in SRL, to demonstrate the typical realization of SRL using model-free approaches (Sec.~\ref{subsec.prel.ppolag}). 
In Sec.~\ref{sec.gensafe}, we will use PPO-Lagrangian as a representative example to explain the integration of the proposed GenSafe with model-free SRL methods.
To better explain our approach, we have provided a list of notations used in this paper in the supplementary material.

\subsection{Constrained Markov Decision Process (CMDP)}
\label{subsec.prel.cmdp}

The CMDP extends the standard MDP by incorporating a constraint function.
A CMDP is defined by the tuple $\mathcal{M}=(S,A,T,R,C,\gamma,\mu)$, where $S$ is the set of states, and $A$ is the set of actions.
$T: S \times A \times S \rightarrow [0,1]$ denotes the state transition kernel, with $T(s,a,s')$ represents the probability of transitioning from state $s\in S$ to state $s' \in S$  under action $a \in A$.
The reward function $R: S\times A \rightarrow \mathbb{R}$ and the cost function $C: S\times A  \rightarrow \mathbb{R}$ assign single-stage reward $r$ and cost $c$ to each state-action pair, respectively.
$\gamma$ is the discount factor and $\mu$ is the initial state distribution. 
In general, both the reward and cost functions in CMDP are considered to be non-negative.

\newtheorem{remark}{Remark}
\begin{remark}
In this work, we consider a single cost function for simplicity.
For scenarios involving multiple cost functions, our proposed method can be extended by constructing a separate ROMDP for each individual cost function. 
Then in the action correction process, the safety constraints can be expanded accordingly using these constructed ROMDPs. 
This can be done by either including each cost function as independent ROMDP-based constraints, or by combining all cost functions into a single expression, e.g., through a weighted combination.
\end{remark}

For a given state $s$, a stationary policy $\pi(s): S \rightarrow A$ determines an action $a$ for the agent to take.
In deep learning contexts, the policy is typically represented by a neural network parameterized by $\theta$, and is denoted as $\pi_\theta$.
Given a policy $\pi_\theta$, the state value function $V_R^{\pi_\theta}(s)$, the state-action value function $Q_R^{\pi_\theta}(s,a)$, and the advantage function $A_R^{\pi_\theta}(s,a)$ for the reward are defined as
\begin{equation}
\label{eq.reward_value_function}
V_R^{\pi_\theta}(s) = \mathbb{E}_{\pi_\theta}\left[ \sum_{t=0}^{\infty} \gamma^t R(s_t,a_t) \middle| s_0 = s \right],
\end{equation}
\begin{equation}
\label{eq.reward_action_function}
Q_R^{\pi_\theta}(s,a) = \mathbb{E}_{\pi_\theta}\left[ \sum_{t=0}^{\infty} \gamma^t R(s_t,a_t) \middle| s_0 = s, a_0 = a \right],
\end{equation}
\begin{equation}
\label{eq.reward_advantage_function}
A_R^{\pi_\theta}(s,a) = Q_R^{\pi_\theta}(s,a) - V_R^{\pi_\theta}(s).
\end{equation}
Similarly, for the cost, the state value function $V_C^{\pi_\theta}(s)$, the state-action value function $Q_C^{\pi_\theta}(s,a)$, and the advantage function $A_C^{\pi_\theta}(s,a)$ are given by substituting the reward function $R$ with the cost function $C$.
For simplicity, the notation $\pi_\theta$ will be omitted in the remainder of this paper when there is no ambiguity.

\subsection{Safe Reinforcement Learning (SRL)}
\label{subsec.prel.srl}
The goal of SRL is to maximize task performance while ensuring the satisfaction of safety requirements. 
Within CMDP, such an SRL problem can be defined through the use of infinite-horizon discounted reward $J_R(\pi_\theta)$ and cost $J_C(\pi_\theta)$ returns, leading to the following optimization problem:
\begin{equation}
\label{eq.srl_optimization}
\max_{\theta} J_R(\pi_\theta), \text{ s.t. } J_C(\pi_\theta) \leq d,
\end{equation}
where $J_R(\pi_\theta) = \mathbb{E}_{\pi_\theta}\left[ \sum_{t=0}^{\infty} \gamma^t R(s_t,a_t) \right]$ and $J_C(\pi_\theta) = \mathbb{E}_{\pi_\theta}\left[ \sum_{t=0}^{\infty} \gamma^t C(s_t,a_t) \right]$.
$d$ is a predefined threshold value.
Hence, the optimal policy $\pi^*_\theta$ of the SRL problem can be determined by solving~\eqref{eq.srl_optimization}. 
Note that, alternative formulations, e.g., finite-horizon undiscounted returns, can also be applied to define the optimization problem \eqref{eq.srl_optimization}.
The proposed GenSafe is adaptable to these different definitions by modifying the reformulated ROMDP-based constraints accordingly, see Sec.~\ref{sec.gensafe} for more details.

\subsection{PPO-Lagrangian}
\label{subsec.prel.ppolag}

Directly solving~\eqref{eq.srl_optimization} is generally infeasible since, on the one hand, both the reward function $R$ and the cost function $C$ are unknown and necessitate approximations.
On the other hand, the constrained nature of the optimization problem poses computational challenges. 
Therefore, a widely employed method in model-free SRL is the introduction of Lagrangian relaxation~\cite{bertsekas2014constrained}. 
By integrating the constraints into the objective function using Lagrange multipliers, Lagrangian relaxation converts a constrained optimization problem into an unconstrained one, simplifying the problem by focusing on optimizing a single combined function.
The Lagrangian of~\eqref{eq.srl_optimization} can be written as
\begin{equation}
L(\theta,\lambda) = J_R(\pi_\theta) - \lambda(J_C(\pi_\theta) - d),
\label{eq.lagrangian}
\end{equation}
where $\lambda \in \mathbb{R}^+$ is the Lagrange multiplier.
It then converts the constrained optimization problem~\eqref{eq.srl_optimization} into the following unconstrained problem
\begin{equation}
L(\theta^*,\lambda^*) = \max_\theta \min_\lambda L(\theta,\lambda),
\label{eq.srl_lag_optimization}
\end{equation}
where the tuple $(\theta^*,\lambda^*)$ indicates the optimal saddle point.
Note that Lagrangian relaxation also faces several challenges, such as weak duality that may lead to suboptimal solutions, the difficulty in selecting appropriate Lagrange multipliers, and the necessity of satisfying specific regularity conditions~\cite{bertsekas2014constrained}.
Nevertheless, it remains a powerful technique for solving constrained optimization problems.

If $J_R(\pi_\theta)$ and $J_C(\pi_\theta)$ are known, \eqref{eq.srl_lag_optimization} can then be solved via gradient search methods~\cite{jayant2022model}.
For this purpose, the PPO-Lagrangian method~\cite{ray2019benchmarking} employs the PPO clipped objectives~\cite{schulman2017proximal} for the estimation of $J_R(\pi_\theta)$ and $J_C(\pi_\theta)$ as
\begin{equation}
J_R(\pi_\theta) = \mathbb{E}_t\left[\min(r_t(\theta)A_R^t, \text{clip}(r_t(\theta), 1-\epsilon, 1+\epsilon)A_R^t) \right],
\label{eq.ppo_reward}
\end{equation}
\begin{equation}
J_C(\pi_\theta) = \mathbb{E}_t\left[\min(r_t(\theta)A_C^t, \text{clip}(r_t(\theta), 1-\epsilon, 1+\epsilon)A_C^t) \right],
\label{eq.ppo_cost}
\end{equation}
where $r_t(\theta) = \frac{\pi_\theta(a_t|s_t)}{\pi_{\theta_{\text{old}}}(a_t|s_t)}$  represents the likelihood ratio of selecting action $a_t$ in state $s_t$ under the parameter $\theta$ compared to $\theta_{\text{old}}$.
$\mathbb{E}_t$ indicates the empirical average over a finite batch of samples, and $\epsilon$ is the clip ratio.
$A_R^t$ and $A_C^t$ are estimated advantages for the reward and cost, which are often computed via GAE as
\begin{equation}
A_R^t = \sum_{l=0}^k (\gamma \bar{\lambda})^l \delta_{R}^{t+l}, \quad A_C^t = \sum_{l=0}^k (\gamma \bar{\lambda})^l \delta_{C}^{t+l}
\label{eq.advantage_estimate} 
\end{equation}
where $\delta_{R}^{t} = r_{t+1} + V_R(s_{t+1}) - V_R(s_t)$ and $\delta_{C}^{t} = c_{t+1} + V_C(s_{t+1}) - V_C(s_t)$ denote the reward and cost temporal differences on the sample path, respectively.
$\bar{\lambda}$ is a parameter that adjusts the bias-variance trade-off in GAE.
In general, $V_R(s)$ and $V_C(s)$ are approximated using neural networks, which are trained with data from sampled trajectories.
However, in the early learning phase, the limited data availability impedes the accuracy of these approximations, leading to diminished safety performance.
To tackle this problem, we therefore propose GenSafe in this work.

\section{Reduced Order Markov Decision Process}
\label{sec.romdp}

The central idea of the proposed GenSafe involves utilizing a Reduced Order Markov Decision Process (ROMDP) as a low-dimensional proxy for approximating the cost function in CMDP. 
In this section, we explain how to construct the ROMDP using order reduction techniques that leverage data observed during the learning process (Section~\ref{subsec.romdp.datasample}). 
The construction process consists of five steps: state abstraction (Section~\ref{subsubsec.romdp.construction.stateabst}), action abstraction (Section~\ref{subsubsec.romdp.construction.actionabst}), cost abstraction (Section~\ref{subsubsec.romdp.construction.costabst}), transition abstraction (Section~\ref{subsubsec.romdp.construction.transabst}), and policy abstraction (Section~\ref{subsubsec.romdp.construction.policyabst}). 
More details are presented as follows.

\subsection{Data Samples}
\label{subsec.romdp.datasample}
During the online learning process, we receive sampled data about the reward and cost returns.
For better explaining the construction of the ROMDP, we first clarify how data samples are defined in this work.

We denote the entire set of currently available data as $\mathcal{D}$, which is defined as
\begin{equation}
\mathcal{D} =\left\{ \begin{array}{c}
    D_1, \\
    D_2, \\
    \ldots \\
    D_i, \\
    \ldots \\
    D_n \\
\end{array} \right\} = \left\{ \begin{array}{c}
    (s_1, a_1, s'_1, r_1, c_1), \\
    (s_2, a_2, s'_2, r_2, c_2), \\
    \ldots \\
    (s_i, a_i, s'_i, r_i, c_i), \\
    \ldots \\
    (s_n, a_n, s'_n, r_n, c_n) \\
\end{array} \right\}
\label{eq.dataset}
\end{equation}
where each data sample represents the information observed in one timestep during the learning process.
$D_i = (s_i,a_i,s'_i,r_i,c_i)$, $i=1,\ldots,n$ represents the $i$-th data sample that contains the current state $s_i$, the applied action $a_i$, the subsequent state $s'_i$ for the given state-action pair, and the observed single-stage reward $r_i$ and cost $c_i$ for this executed state-action pair.
The size of the dataset is denoted by $|\mathcal{D}| = n$.

Accordingly, we can group the information from all data samples to obtain the set of all current states included in $\mathcal{D}$ as $\mathcal{D}_{s} = \{s_1, \ldots, s_i,\ldots, s_n \}$, where each state $s_i$ corresponds to the information contained in the data sample $D_i$.
Similarly, we use $\mathcal{D}_{a}  = \{a_1, \ldots, a_n \}$,  $\mathcal{D}_{s'}  = \{s'_1, \ldots, s'_n \}$, and $\mathcal{D}_{c}  = \{c_1, \ldots, c_n \}$ to represent the set of applied actions, subsequent states, and costs in $\mathcal{D}$, respectively.
These datasets can be viewed as categorizing different aspects of the information contained in $\mathcal{D}$.
Since they are all extracted from $\mathcal{D}$ using each data sample $D_i$, we have $|\mathcal{D}_{s}|=|\mathcal{D}_{a}|=|\mathcal{D}_{s'}|=|\mathcal{D}_{c}|=|\mathcal{D}|=n$.
We then construct a low-dimensional ROMDP by utilizing these datasets.


\subsection{Construction of ROMDP}
\label{subsec.romdp.construction}

In complex autonomous systems, the state $s$ usually comprises observations of numerous physical attributes, resulting in a high-dimensional state space $S \subseteq \mathbb{R}^{n_s}$.
This high dimensionality presents computational challenges for function approximation, especially when data samples are insufficient.
To address this issue, we employ order reduction techniques to derive a representative low-dimensional safety predictor model, i.e., the ROMDP, to estimate the safety of the original system.
This is achieved by first transforming the original state $s$ into a new, simplified state $s^r$ (referred to as the reduced state in this paper) with a reduced dimensionality of the state space $S^r \subseteq \mathbb{R}^{n_{S^r}}$, i.e., $n_{S^r} \ll n_s$.
Following this transformation, we construct a ROMDP based on the reduced state $s^r$ accordingly, which is defined as follows.

\begin{figure}
    \centering
    \includegraphics[trim={5mm 15mm 5mm 12mm}, width=.9\linewidth]{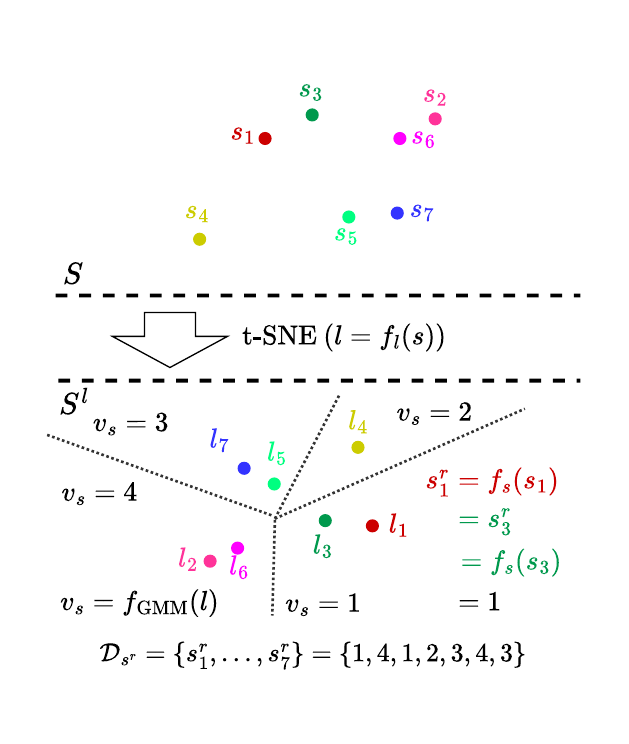}
    \caption{Example of the state abstraction. Through applying t-SNE, a set of original states $\mathcal{D}_s =\{s_1, \ldots, s_7\}$ is transformed into a corresponding set of deterministic low-dimensional states $\mathcal{D}_l = \{l_1, \ldots, l_7\}$, where similar high-dimensional data points are represented by nearby low-dimensional states. 
    The mapping function $f_l$, trained with $\mathcal{D}_s$ and $\mathcal{D}_l$, reduces the high-dimensional state space $S$ to a two-dimensional state space $S^l$. 
    The GMM classifier $f_{\text{GMM}}$ then divides $S^l$ into $k_s=4$ regions, each assigned with an index $v_s \in \{1,2,3,4\}$.
    The reduced state $s^r$ is thus determined by using the state abstraction function $s^r = f_s(s) = f_{\text{GMM}}(f_l(s))$, e.g., we have $s^r_1 = f_s(s_1) = s^r_3 = f_s(s_3) = 1$ and $\mathcal{D}_{s^r} = \{s^r_1,\ldots,s^r_7\} = \{1,4,1,2,3,4,3\}$ in this example.}
    \label{fig.tsne}
    \vspace{-10pt}
\end{figure}

\newtheorem{definition}{Definition}
\begin{definition}[ROMDP]
Given a CMDP $\mathcal{M} = (S,A,T,R,C,\gamma,\mu)$, a corresponding ROMDP $\mathcal{M}^r = (S^r, A^r, T^r, C^r, \gamma, \mu^r)$ is constructed by using the reduced state $s^r \in  S^r$ and the reduced action $a^r \in A^r$, where $s^r = f_s(s)$ and $a^r = f_a(a)$. 
$f_s: S \rightarrow S^r$ and $f_a: A \rightarrow A^r$ are referred to as the state abstraction function and the action abstraction function, respectively.
$T^r: S^r \times A^r \times S^r \rightarrow [0,1]$ is the reduced transition kernel for the reduced states and actions.
$C^r: S^r \times A^r \rightarrow \mathbb{R}$ denotes the reduced cost function that serves as a low-dimensional approximation of $C$.
The discount factor $\gamma$ remains the same as in the CMDP.
The initial state distribution $\mu^r$ is derived by applying $f_s$ to $\mu$ accordingly.
\end{definition}

Since the ROMDP is specifically employed to improve constraint satisfaction in SRL, it incorporates only the cost function. 
By assuming that high-dimensional state and action pairs mapping to the same low-dimensional pairs also exhibit similar safety characteristics, the ROMDP is able to efficiently predict actual cost returns for unobserved control commands based on limited data.
In the following, we present details about the construction of ROMDP, which consists of state abstraction, action abstraction, cost abstraction, transition abstraction and policy abstraction.

\subsubsection{State Abstraction}
\label{subsubsec.romdp.construction.stateabst}
To determine the state abstraction function $f_s$, we first apply an order reduction technique to transform the set of original states $\mathcal{D}_{s} = \{s_1, \ldots, s_n \}$ into a corresponding set of low-dimensional states. 
For this, we propose using t-distributed Stochastic Neighbor Embedding (t-SNE)~\cite{van2008visualizing}, a method originally developed for visualizing high-dimensional data. 
Compared to traditional methods like Principal Component Analysis (PCA)~\cite{abdi2010principal}, t-SNE is able to more effectively distinguish and classify high-dimensional data.
Through the usage of Euclidean distance as the similarity metric, t-SNE projects high-dimensional data points into a two- or three-dimensional state space $S^l$, in which similar high-dimensional data points are represented by nearby low-dimensional data points (see Fig.~\ref{fig.tsne} for an example).
In this work, we consider the order reduction to a two-dimensional state space, i.e., $S^l \subseteq \mathbb{R}^2$.
Meanwhile, to optimize the performance of t-SNE, we normalize the magnitudes of each state component to the same range.
The result of applying t-SNE is a set of deterministic low-dimensional states $\mathcal{D}_{l} = \{l_1, \ldots, l_n\}$, where each state $l_i$ is the reduced representation of its original state $s_i$. 
For a more comprehensive explanation of how t-SNE computes these low-dimensional states, please refer to the supplementary material of this paper and~\cite{van2008visualizing}.

Note that, while t-SNE effectively determines the values of the low-dimensional states, it does not provide an explicit expression for the mapping function $l = f_l(s): S \rightarrow S^l$.
To address this, we train a neural network to approximate this mapping function using the set of original states $\mathcal{D}_s$ and the corresponding set of low-dimensional states $\mathcal{D}_l$.
Then, by leveraging the mapping function $f_l$, we are able to reduce the dimensionality of the original state space $S$ to a two-dimensional state space $S^l$. 
However, dealing with a continuous state space often still presents computational challenges. 
Therefore, to further improve computational efficiency and make it more manageable for cost function approximation, we proceed to discretize the low-dimensional state space $S^l$.

A frequently used approach for discretization is dividing the entire state space $S^l$ into grids of fixed sizes. 
However, this method may not be effective when data points are unevenly distributed across the state space (see Fig.~\ref{fig.low_states} for an example). 
Therefore, in this work, we employ a Gaussian Mixture Model (GMM)~\cite{reynolds2009gaussian,mclachlan1988mixture} classifier, which is trained on $\mathcal{D}_l$, for discretization.
With a predefined number of clusters $k_s$, the GMM classifier segments the low-dimensional state space $S^l$ into $k_s$ distinct cluster regions, with each region being assigned an index $v_s \in \{1,2,\ldots,k_s\}$.
For any given low-dimensional state $l$, its corresponding cluster index is determined by $v_s = f_\text{GMM}(l)$, where $f_\text{GMM}: S^l \rightarrow \{1,2,\ldots,k_s\}$ denotes the GMM-based classification process. 
Hence, the reduced state $s^r$ can be effectively represented by the cluster index $v_s$, and the reduced state space $S^r$ becomes the collection of all these cluster indices, i.e., we have $s^r = v_s \in  S^r = \{1,\ldots,k_s\}$.
Consequently, the state abstraction function $f_s$ is defined as
\begin{equation}
s^r = f_s(s) = f_\text{GMM}(f_l(s)),
\label{eq.state_abstraction}
\end{equation}
which indicates that the original state $s$ is first transformed into a corresponding low-dimensional state $l$, and then is classified by the GMM classifier (see Fig.~\ref{fig.tsne}).
We denote the set of reduced states that corresponds to $\mathcal{D}_s$ as $\mathcal{D}_{s^r} = \{s^r_1,\ldots,s^r_n\}$.

\subsubsection{Action Abstraction}
\label{subsubsec.romdp.construction.actionabst}
\begin{figure}
    \centering
    \includegraphics[trim={5mm 8mm 7mm 5mm}, width=.9\linewidth]{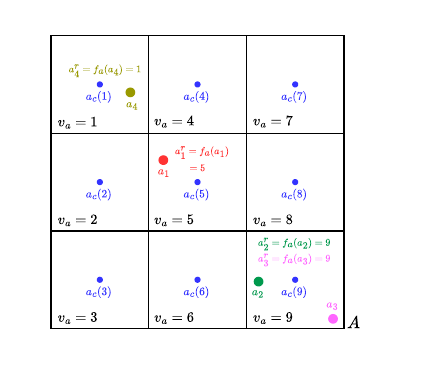}
    \caption{Example of the action abstraction. For a two-dimensional action space $A \subseteq \mathbb{R}^2$, we discretize it using $k_a =3$, which results in a total of $k_a^{n_a}=9$ grids. Each grid cell is assigned an index $v_a \in \{1,\ldots,9\}$. $a_c(1),\ldots,a_c(9)$ denote the center of each grid. For a set of original applied actions $\mathcal{D}_a = \{a_1,\ldots,a_4 \}$, we thus have $a^r_1 = f_a(a_1) =5$, $a^r_2 = f_a(a_2) = a^r_3 = f_a(a_3) = 9$, $a^r_4 = f_a(a_4) = 1$. }
    \label{fig.action_abstraction}
    \vspace{-10pt}
\end{figure}

Normally, the action space $A \subseteq \mathbb{R}^{n_a}$ has a significantly lower dimensionality than the state space $S$. 
Inspired by this, we adopt a straightforward approach for the action abstraction by directly discretizing the action space $A$ into uniformly sized grid cells (see Fig.~\ref{fig.action_abstraction} for an example).
Compared to the use of t-SNE and the GMM classifier, such an approach is able to enhance computational efficiency and reduce approximation errors inherent in the order reduction process.
Given a specified number of grids per dimension $k_a$, the direct discretization therefore results in a total of $k_a^{n_a}$ grids.
We assign each of these grids an index $v_a \in \{1,2,\ldots,k_a^{n_a}\}$.
In a manner analogous to the reduced states, the reduced action $a^r$ is thus represented by this index $v_a$, and the reduced action space $A^r$ is the collection of all such indices, i.e., $a^r = v_a \in A^r = \{1,2, \ldots,k_a^{n_a}\}$.
The action abstraction function $f_a$ is thus defined as
\begin{equation}
a^r = f_a(a) = \argmin_{v_a} || a - a_c(v_a) ||^2
\label{eq.action_abstraction}
\end{equation}
where $a_c(v_a)$ indicates the center of the grid cell indexed by $v_a$ (see Fig.~\ref{fig.action_abstraction} for an example).
$f_a$ can be considered as a localization function that, for a given action $a$, determines the specific grid cell it falls within and then returns the corresponding index $v_a$. 
The set of reduced actions that corresponds to $\mathcal{D}_a$ are denoted as $\mathcal{D}_{a^r} = \{a^r_1,\ldots,a^r_n\}$.

\begin{remark}
The prerequisite for directly discretizing the action space $A$ is that it should has a relatively small dimensionality, thereby making the process computationally manageable.
For instance, in this work, we assume $n_a = 2$ or $n_a = 3$. 
In cases where the action space has a higher dimensionality, rendering the direct discretization impractical, a similar order reduction technique used for the state abstraction can be employed. 
This involves first using t-SNE to identify a low-dimensional action space, and then applying a GMM classifier to determine the corresponding reduced actions.
\end{remark}

\subsubsection{Cost Abstraction}
\label{subsubsec.romdp.construction.costabst}
After the state and action abstractions, we are now able to approximate the original cost function $C$ with a reduced cost function $C^r$, which leverages the reduced states and actions. 
To achieve this, we first reorganize the set of observed costs $\mathcal{D}_c = \{c_1, \ldots, c_n \}$ by using the derived sets of reduced states $\mathcal{D}_{s^r}$ and reduced actions $\mathcal{D}_{a^r}$.
Note that, the elements within these datasets that share the same index $i$ (i.e., $c_i$, $s^r_i$, and $a^r_i$) all correspond to the same data sample $D_i$, representing different facets of the information it contains.
Hence, for each specific pair of the reduced state and action, we can sort all data samples in $\mathcal{D}$ to identify a corresponding set of observed costs as
\begin{equation}
\mathcal{D}^C_{s^r,a^r} = \{c_i | f_s(s_i) = s^r, f_a(a_i) = a^r \},
\label{eq.cost_sets}
\end{equation}
which includes all costs whose corresponding data samples have the specified reduced state $s^r$ and action $a^r$.
The size of this dataset is denoted as $|\mathcal{D}^C_{s^r,a^r}| = n_{s^r,a^r}$.
Note that, $ n_{s^r,a^r}$ also represents the number of data samples in $\mathcal{D}$ that, after transforming the original state and action to their reduced forms, correspond to the specified reduced state $s^r$ and action $a^r$, i.e., $n_{s^r,a^r} = |\{ D_i \in \mathcal{D} | f_s(s_i) = s^r, f_a(a_i) = a^r \}|$. 

Then, the reduced cost function $C^r$ is computed as follows
\begin{equation}
C^r(s^r, a^r) = \begin{cases}
 {\displaystyle \frac{\sum_{c_i \in \mathcal{D}^C_{s^r,a^r}} c_i}{n_{s^r,a^r}}, }  & \text{ if } n_{s^r,a^r} > 0, \\
{\displaystyle \Delta }, & \text{ if } n_{s^r,a^r} = 0,
\end{cases}
\label{eq.reduced_cost_function}
\end{equation}
which indicates that the reduced cost is determined by averaging the observed costs within $\mathcal{D}^C_{s^r,a^r}$ if this set is not empty.
When no observed costs are available for a given reduced state and action pair, the reduced cost is set to a predefined estimated value $\Delta$.
Thus, the value of the original cost function $C(s,a)$ can be approximated by using the reduced cost function $C^r$ through converting the original state $s$ and action $a$ into their corresponding reduced state $s^r$ and action $a^r$, i.e., $C(s,a) \approx C^r(s^r, a^r)$ with $s^r = f_s(s)$ and $a^r = f_a(a)$ (see also Example~\ref{example_cost}). 
Consequently, the reduced cost function $C^r$ serves as a low-dimensional approximator of the original cost function $C$, enhancing the efficiency of estimation and computation processes.

\newtheorem{example}{Example}
\begin{example}
\label{example_cost}
Consider the entire dataset $\mathcal{D} =\{D_1, \ldots, D_5\}$ contains 5 data samples, the corresponding sets of original states $\mathcal{D}_s = \{ s_1,\ldots, s_5 \}$, original actions $\mathcal{D}_a = \{ a_1,\ldots, a_5 \}$ and observed costs $\mathcal{D}_c = \{ c_1,\ldots, c_5 \}$ therefore also have 5 elements.
Suppose that after applying the state and action abstraction functions, we obtain the corresponding sets of reduced states and actions as $\mathcal{D}_{s^r} = \{s^r_1,\ldots,s^r_5\} = \{1,1,2,2,3\}$ and $\mathcal{D}_{a^r} = \{a^r_1,\ldots,a^r_5\} = \{1,1,1,2,2\}$, where we assume that $S^r = \{1,2,3\}$ and $A^r = \{1,2\}$.
Thus, by sorting the dataset $\mathcal{D}$, we identify the following sets of observed costs $\mathcal{D}^C_{s^r,a^r}$ for each reduced state and action pair:
\begin{alignat}{2}
 &\mathcal{D}^C_{1,1} = \{c_1, c_2\}, \quad & \mathcal{D}^C_{1,2} &= \varnothing , \nonumber \\
 &\mathcal{D}^C_{2,1} = \{c_3\}, & \mathcal{D}^C_{2,2}  & = \{c_4 \}, \nonumber \\
 & \mathcal{D}^C_{3,1} = \varnothing, & \mathcal{D}^C_{3,2} & = \{ c_5\},
\end{alignat}
with $n_{1,1} = |\{D_1, D_2 \}|= 2$,  $n_{1,2} = |\varnothing| = 0$, $n_{2,1} = |\{D_3 \}| = 1$, $n_{2,2} = |\{D_4 \}| =1$, $n_{3,1} = |\varnothing| = 0$, $n_{3,2} = |\{D_5 \}| = 1$.
Therefore, the reduced cost function is calculated as 
\begin{alignat}{2}
 &C^r(1, 1)  = \frac{c_1 + c_2}{2}, \quad & C^r(1, 2)  = \Delta , \nonumber \\
 &C^r(2, 1)  = c_3, & C^r(2, 2)  = c_4, \nonumber \\
 &C^r(3, 1)  = \Delta, & C^r(3, 2)  = c_5. 
\end{alignat}
For any new state and action pair, the original cost function $C(s,a)$ is estimated by using the corresponding reduced state and action, e.g., if $f_s(s_6) = s^r_6 = 2$ and $f_a(a_6) = a^r_6 = 1$, then we have $C(s_6, a_6) \approx C^r(s^r_6,  a^r_6) = C^r(2,1) = c_3$.
\end{example}

Note that, while the single-stage cost can be directly estimated using the reduced cost function $C^r$, solving the SRL problem~\eqref{eq.srl_optimization} often requires the estimation of future costs $J_C(\pi_\theta)$ under the current policy $\pi_\theta$.
Hence, to facilitate an accurate estimation of these future costs, we further incorporate transition abstraction and policy abstraction into the construction of ROMDP. 

\begin{remark}
In general, the number of reduced states $k_s$ can be determined by balancing the representational capability of the reduced states and the number of available data samples. 
This can be achieved by first estimating the number of data samples required for each reduced state to be sufficiently representative, and then dividing the total number of available data samples by this estimated number. 
Meanwhile, the number of reduced actions $k_a^{n_a}$ is suggested to be chosen within a similar range as the number of reduced states.
The initial estimated cost $\Delta$ for unobserved reduced state and action pairs can be set at approximately half of the maximum single-stage cost defined in the environmental setup. 
This provides a relatively neutral estimation for unknown actions.
\end{remark}

\subsubsection{Transition Abstraction}
\label{subsubsec.romdp.construction.transabst}
To determine the reduced transition kernel $T^r$, we first transform the set of observed subsequent states $\mathcal{D}_{s'}  = \{s'_1, \ldots, s'_n \}$ into a corresponding set of subsequent reduced states $\mathcal{D}_{s^{r'}}  = \{s^{r'}_1, \ldots, s^{r'}_n \}$ using the state abstraction function $f_s$.
Similar to the process of cost abstraction, we then identify a corresponding set of subsequent reduced states for each reduced state and action pair by reorganizing  $\mathcal{D}_{s^{r'}}$ alongside its associated $\mathcal{D}_{s^r}$ and  $\mathcal{D}_{a^r}$ 
\begin{equation}
\mathcal{D}^{s^{r'}}_{s^r,a^r} = \{s^{r'}_i | f_s(s_i) = s^r, f_a(a_i) = a^r  \},
\label{eq.subset_state_set}
\end{equation}
where each element represents an observed data sample for the transition path $s^r\times a^r \rightarrow s^{r'}$.
Note that, the size of this dataset is the same as $\mathcal{D}^C_{s^r,a^r}$, i.e., $|\mathcal{D}^{s^{r'}}_{s^r,a^r}| = n_{s^r,a^r}$, as both datasets are derived by sorting the entire dataset $\mathcal{D}$ according to the reduced state and action pairs. 
Subsequently, we count the number of observed data samples for each possible transition path $s^r\times a^r \rightarrow s^{r'}$, which is denoted as $n_{s^r\times a^r \rightarrow s^{r'}}$.
The reduced transition kernel $T^r$ is thus computed as  
\begin{equation}
T^r(s^r, a^r, s^{r'}) = \begin{cases}
 \frac{n_{s^r\times a^r \rightarrow s^{r'}}}{n_{s^r,a^r}}    , & \text{if } n_{s^r,a^r} > 0, \\
\frac{1}{k_s}, & \text{if } n_{s^r,a^r} = 0, 
\end{cases}
\label{eq.reduced_transition_kernel}
\end{equation}
where $k_s = |S^r|$ is the size of the reduced state space.  
This implies that when data samples are available for a specific reduced state and action pair, the probability of transitioning to a particular subsequent reduced state $s^{r'}$ is proportional to the observed frequency of this transition path $s^r\times a^r \rightarrow s^{r'}$ among all observed data samples for this pair (see Example~\ref{example_transition}).
Conversely, if no data exists, all subsequent reduced states are assigned equal transition probabilities.
Hence, for each given pair of the reduced state $s^r$ and action $a^r$, we ensure that $\sum_{s^{r'} \in S^r} T^r(s^r, a^r, s^{r'}) = 1$ is satisfied.

\begin{example}
\label{example_transition}    
Continued from Example~\ref{example_cost}, we assume that the corresponding set of subsequent reduced states is $\mathcal{D}_{s^{r'}}  = \{s^{r'}_1, \ldots, s^{r'}_5 \} = \{1,2,2,3,1 \}$.
Hence, we have 
\begin{alignat}{2}
 &\mathcal{D}^{s^{r'}}_{1,1} = \{s^{r'}_1, s^{r'}_2\} = \{1,2\}, \quad & \mathcal{D}^{s^{r'}}_{1,2} &= \varnothing , \nonumber \\
 &\mathcal{D}^{s^{r'}}_{2,1} = \{s^{r'}_3\} = \{ 2 \}, & \mathcal{D}^{s^{r'}}_{2,2}  & = \{s^{r'}_4 \} = \{ 3 \}, \nonumber \\
 & \mathcal{D}^{s^{r'}}_{3,1} = \varnothing, & \mathcal{D}^{s^{r'}}_{3,2} & = \{ s^{r'}_5\}  = \{1 \}, 
\end{alignat}
with the counts of observed transition paths as $n_{1\times 1 \rightarrow 1} = 1$, $n_{1\times 1 \rightarrow 2} = 1$, $n_{2\times 1 \rightarrow 2} = 1$, $n_{2\times 2 \rightarrow 3} = 1$, $n_{3\times 2 \rightarrow 1} = 1$, and $n_{s^r\times a^r \rightarrow s^{r'}} = 0$ for any other transition paths.
The reduced transition kernel $T^r$ is thus calculated by using these counts.
For instance, since $n_{1,1} = 2$, $n_{1,2} = 0$, $n_{2,1} = 1$, we have 
\begin{alignat}{3}
 & T^r(1, 1, 1)  = \frac{1}{2}, \quad & T^r(1, 1, 2)  & = \frac{1}{2} , \quad  & T^r(1, 1, 3)  = 0, \nonumber \\
 & T^r(1, 2, 1)  = \frac{1}{3}, \quad & T^r(1, 2, 2) & = \frac{1}{3} , \quad  & T^r(1, 2, 3)  = \frac{1}{3}, \nonumber \\
 & T^r(2, 1, 1)  = 0, \quad & T^r(2, 1, 2) & = 1, \quad  & T^r(2, 1, 3)  = 0.
\end{alignat}

\end{example}

\subsubsection{Policy Abstraction} 
\label{subsubsec.romdp.construction.policyabst}
To predict future costs, we also construct a reduced policy $\pi^r (s^r): S^r \rightarrow A^r$ that emulates the original policy $\pi_\theta$. 
It determines the probability of selecting a specific reduced action $a^r$ for a given reduced state $s^r$.
Note that, since the cost function $C(s,a)$ and transition kernel $T$ are independent of the control policy, we are able to use all data collected by different policies, i.e., the entire dataset $\mathcal{D}$, for the cost and transition abstractions. 
However, the control policy $\pi_\theta$ is updated after each learning epoch. 
As a result, only the data collected in the last epoch represents the up-to-date control policy $\pi_\theta$ and can be used for an accurate policy abstraction.
Therefore, we derive a sub-dataset $\mathcal{D}^{e}$ from the entire dataset $\mathcal{D}$, which contains only the data samples collected in the last epoch, for the policy abstraction.

For each unique pair of reduced state $s^r$ and action $a^r$, we first count the number of data samples $n^e_{s^r,a^r}$ within the dataset $\mathcal{D}^{e}$ that correspond to this pair.
These counts reflect instances where, upon converting the original state and action to their reduced forms, the original policy $\pi_\theta$ selects $a^r$ as the action for the reduced state $s^r$. 
The total number of observed pairs for a specific reduced state $s^r$ within $\mathcal{D}^{e}$ is denoted as $n^e_{s^r}$, where we have $n^e_{s^r} = \sum_{a^r \in A^r} n^e_{s^r,a^r}$.
Then, the reduced policy $\pi^r$ is defined by the following probabilities
\begin{equation}
\mathbb{P}( \pi^r(s^r) = a^r) = \begin{cases}
 \frac{{n^e_{s^r,a^r}}}{{n^e_{s^r}}}    , & \text{if } n_{s^r} > 0, \\
\frac{1}{k_a^{n_a}}, & \text{if } n_{s^r} = 0, 
\end{cases}
\label{eq.reduced_policy}
\end{equation}
where $k_a^{n_a} = |A^r|$ is the size of the reduced action space.
Similarly, this indicates that when data are available, the probability of choosing the reduced action $a^r$ for the reduced state $s^r$ is proportional to its observed frequency relative to the total observations for this state (see Example~\ref{example_reduced_policy}). 
For reduced states $s^r$ without data, we adopt an equal probability distribution across all possible reduced actions.
Therefore, for each given reduced state $s^r$, we have $\sum_{a^r \in A^r}\mathbb{P}( \pi^r(s^r) = a^r) = 1$.

\begin{example}
\label{example_reduced_policy}
Based on Example~\ref{example_cost}, we assume that $\mathcal{D}^{e} = \mathcal{D}$. 
Hence, we have the following counts for each reduced state and action pair: $n^e_{1,1} = 2$,  $n^e_{1,2} = 0$, $n^e_{2,1} = 1$, $n^e_{2,2} = 1$, $n^e_{3,1} = 0$, $n^e_{3,2} = 1$.
Accordingly, we have $n^e_1 = n^e_{1,1} + n^e_{1,2} = 2$, $n^e_2 = n^e_{2,1} + n^e_{2,2} = 2$, $n^e_3 = n^e_{3,1} + n^e_{3,2} = 1$.
The reduced policy $\pi^r$ is thus represented by the following probabilities
\begin{alignat}{2}
 &\mathbb{P}( \pi^r(1) = 1) = 1, \quad  &\mathbb{P}( \pi^r(1) = 2) & = 0, \nonumber \\
 & \mathbb{P}( \pi^r(2) = 1) = \frac{1}{2}, &\mathbb{P}( \pi^r(2) = 2) & = \frac{1}{2}, \nonumber \\
 & \mathbb{P}( \pi^r(3) = 1) = 0, &\mathbb{P}( \pi^r(3) = 2) &= 1. 
\end{alignat}
\end{example}

Through the aforementioned comprehensive abstraction processes, we develop a ROMDP model that functions as a low-dimensional approximator of the cost function in CMDP. 
It is capable of efficiently predicting not only the single-stage costs but also the future costs under a given policy $\pi_\theta$.
In the next section, we explain how to integrate this constructed ROMDP into our proposed GenSafe for improving the safety performance of SRL algorithms.

\section{Generalizable Safety Enhancer}
\label{sec.gensafe}

The fundamental concept behind the proposed Generalizable Safety Enhancer (GenSafe) is to reformulate the original cost constraint in CMDP into constraints based on the ROMDP. 
Then, GenSafe enhances the performance of SRL algorithms by modifying each applied action in accordance with these revised constraints. 
In essence, GenSafe acts as an additional safety layer for SRL algorithms (see Fig.~\ref{fig.overview}).
Further details are provided in the remaining part of this section.

\subsection{Action Correction Based on ROMDP}

Solving the SRL problem~\eqref{eq.srl_optimization} necessitates the prediction of future costs $J_C(\pi_\theta)$ under a given policy $\pi_\theta$.
However, accurately estimating $J_C(\pi_\theta)$ is a challenging task and often demands complex computations, e.g., \eqref{eq.ppo_cost}-\eqref{eq.advantage_estimate} in PPO-Lagrangian.
To enable a more reliable and efficient SRL process during the early learning phase, we utilize the constructed ROMDP to approximate cost returns. 
Nonetheless, it should be noted that a certain degree of information loss is inevitable in the order reduction process, resulting in discrepancies between the estimated and the actual values.
Such inaccuracies can accumulate over time, especially with longer prediction horizons. 
This complicates the task of performing a reliable policy-level modification, i.e., directly revising the policy $\pi_\theta$, based on ROMDP predictions.

To mitigate this issue, we propose an action-level correction mechanism that focuses on the short-term outcomes of actions. 
By adjusting only the current action instead of the entire policy, such a mechanism is able to fully utilize the more accurate short-term prediction capabilities of the ROMDP and effectively reduce the need for long-horizon prediction accuracy.
At each timestep $t$ of the learning process, the current SRL policy proposes an action $a_t = \pi_\theta(s_t)$ based on the current state $s_t$.
Then, we solve the following optimization problem to refine this action $a_t$ such that the modified action $a_m$ is more likely to satisfy the safety constraints
\begin{eqnarray}
a_m &=& \argmin_{a} || a - a_t ||^2,
\label{eq.action_correction} \\
 & \text{s.t. }& C^r(s^r_t,a^r) \leq d_s ,
\label{eq.single_constraint} \\
 && C^r(s^r_t,a^r) + \gamma V^r_{C^r}(s^{r'}_t) \leq d ,
\label{eq.future_constraint}
\end{eqnarray}
where $s^r_t = f_s(s_t)$ and $a^r = f_a(a)$ are the corresponding reduced state and action, respectively.
$s^{r'}_t$ denotes the subsequent reduced state resulting from applying $a^r$ to $s^r_t$, i.e., $s^r_t \times a^r \rightarrow s^{r'}_t$.
The goal of this optimization is to find an action $a_m$ that remains as close as possible to the initially determined action $a_t$ while satisfying the constraints \eqref{eq.single_constraint} and \eqref{eq.future_constraint}. 
An action meeting these criteria is then considered safe for the current state $s_t$.
\eqref{eq.single_constraint} imposes an immediate cost constraint that the single-stage cost estimated by the ROMDP should not exceed a predefined threshold $d_s$.
This constraint is motivated by the fact that, in various autonomous systems, e.g., autonomous vehicles or robotic manipulators, the immediate cost can significantly impact system safety~\cite{dalal2018safe}.
Therefore, we additionally incorporate this constraint into the optimization process.
\eqref{eq.future_constraint} mirrors the cost constraint $V_C(s_t) = C(s_t,a) + \gamma  V_C (s_t^{'})\leq d$ for the current state $s_t$, where $s_t \times a \rightarrow s_t^{'}$ and the state, action and cost are substituted with their reduced forms contained in the ROMDP, respectively.
$V^r_{C^r}(s^r)$ denotes the reduced value function for the reduced state $s^r$ and cost $C^r$ under the reduced policy $\pi^r$.
Taking into account the reduced transition kernel $T^r$, \eqref{eq.future_constraint} can be further reformulated as
\begin{equation}
C^r(s^r_t,a^r) + \gamma \sum_{s^r \in S^r}  T^r(s^r_t, a^r, s^r) V^r_{C^r}(s^{r}) \leq d.
\label{eq.future_constraint_revised}
\end{equation}

\begin{remark}
In practice, $d_s$ can be selected by first estimating a reasonable horizon for approximating the cost value function $V_C(s_t)$ based on the actual task requirements. 
Then, $d_s$ can be calculated by dividing $d$ by this horizon. 
Meanwhile, it is advisable to select $d_s$ to be smaller than the estimated cost $\Delta$ for unobserved reduced state and action pairs. 
This will discourage the exploration of previously unknown actions, thereby improving the safety performance.
\end{remark}

Note that, although the reduced cost $C^r$ is identified during the ROMDP construction, tackling~\eqref{eq.future_constraint_revised} requires the computation of the reduced value function $V^r_{C^r}(s^r)$ for each reduced state $s^r$.
For this, we adapt the value iteration algorithm~\cite{Sutton1998RL} by replacing the traditional maximum operation with an average operation that accounts for the probabilities defined by the reduced policy $\pi^r$ (see Algorithm~\ref{alg.value_iteration}).
After initializing the values of all reduced states to zero, we iteratively update $V^r_{C^r}(s^r)$ for each reduced state $s^{r,i}$, $i=1,\ldots,k_s$ contained in the reduced state space $S^r$ by utilizing the Bellman equation
\begin{equation}
V^r_{C^r}(s^{r,i}) = \sum_{a^r \in A^r} \mathbb{P}( \pi^r(s^{r,i}) = a^r) V^r_{s^{r,i},a^r},
\label{eq.value_iteration_update}
\end{equation}
with 
\begin{equation}
\hspace{-1mm} V^r_{s^{r,i},a^r} = C^r(s^{r,i}, a^r) + \gamma \sum_{s^r \in S^r}  T^r(s^{r,i}, a^r, s^r) V^r_{C^r}(s^{r}),
\label{eq.value_iteartion_single}
\end{equation}
where the values are calculated through averaging the probabilities given by the reduced transition kernel $T^r$ and policy $\pi^r$, rather than the maximum operation employed in standard value iteration. 
The resulting reduced value function $V^r_{C^r}(s^r)$ estimates the future cost returns for any given reduced state $s^r$ under the reduced policy $\pi^r$.
Same as the standard value iteration for MDPs, the update is terminated when the difference in the value function between two successive iterations falls beneath a predefined threshold $\delta$.
This ensures efficient convergence of the value function while maintaining a balance between computational cost and precision.

Once the reduced value function $V^r_{C^r}(s^r)$ is determined, verifying constraints \eqref{eq.single_constraint} and \eqref{eq.future_constraint_revised} for any chosen action $a$ becomes straightforward, akin to referencing a look-up table.
However, due to the non-differentiable nature of both the reduced cost $C^r$ and value function $V^r_{C^r}(s^r)$, solving the optimization problem~\eqref{eq.action_correction} through gradient-based methods is not feasible. 
Therefore, we employ derivative-free optimization techniques to find the modified action $a_m$.
Specifically, we use Particle Swarm Optimization (PSO)~\cite{kennedy95particle} in this work.
Consequently, the proposed GenSafe incorporates this action correction mechanism, grounded in ROMDP predictions, to enhance the safety performance of SRL algorithms.

\begin{algorithm}[t]
\caption{Value iteration for calculating $V^r_{C^r}(s^r)$ }
\label{alg.value_iteration}
\begin{algorithmic}[1]
\small
\Require The constructed ROMDP, tolerance $\delta$
\State Initialize $V^r_{C^r}(s^r) = 0$ for all $s^r \in S^r$
\State $\Delta_V = \delta$ 
\While{$\Delta_V \geq \delta$}
\State $\Delta_V = 0$ 
\For{$s^{r,i} \in S^r$, $i = 1,\ldots,k_s$}
    \State $V_{\mathbf{current}} = V^r_{C^r}(s^{r,i})$
    \State Update $V^r_{C^r}(s^{r,i})$ according to~\eqref{eq.value_iteration_update}-\eqref{eq.value_iteartion_single}
    \State $\Delta_V = \max (\Delta_V , |V_{\mathbf{current}} - V^r_{C^r}(s^{r,i})|)$
\EndFor
\EndWhile
\end{algorithmic}
\end{algorithm}

\subsection{SRL with GenSafe}

The integration of the proposed GenSafe with a chosen model-free SRL algorithm, e.g., PPO-Lagrangian, is delineated in Algorithm~\ref{alg.srl_gensafe}.
Throughout the learning process, at each timestep, the SRL policy $\pi_\theta$ determines an action $a_t$ based on the current state $s_t$.
Subsequently, provided that the ROMDP has been constructed, i.e., except during the initial epoch, the proposed GenSafe solves the optimization problem~\eqref{eq.action_correction} to identify a modified action $a_m$.
This action is then applied to the system, with the resultant observed data sample being incorporated into the datasets $\mathcal{D}$ and $\mathcal{D}^e$. 
Upon reaching a predefined number of timesteps within each epoch, the ROMDP is reconstructed leveraging the most recent datasets $\mathcal{D}$ and $\mathcal{D}^e$.  
Meanwhile, the SRL policy $\pi_\theta$ is updated according to the selected SRL algorithm.
In essence, GenSafe acts as an additional safety layer to SRL algorithms (see also Fig.~\ref{fig.overview}).
Given its focus on action-level corrections, GenSafe is inherently compatible with a broad spectrum of model-free SRL algorithms, enhancing their safety performance without necessitating modifications to their core methodologies.

\begin{algorithm}[t]
\caption{SRL with GenSafe}
\label{alg.srl_gensafe}
\begin{algorithmic}[1]
\small
\Require Number of training epochs $k_{e}$, number of timesteps per each epoch $k_{t}$, deactivate threshold $\delta_d$, reactivate threshold $\delta_r$
\State Initialize SRL policy $\pi_\theta$, ROMDP $\mathcal{M}^r = \varnothing$
\State $i = 0$, $\mathcal{D} = \varnothing$, $\mathcal{D}^e = \varnothing$, $\mathcal{M}^r$ is active
\While{$i < k_e$} \Comment{Training loop}
\State $j = 0$
\While{$j < k_t$} \Comment{Collect data}
    \State $a_t = \pi_\theta(s_t)$ \Comment{SRL policy}
    \If{$\mathcal{M}^r \neq \varnothing$ and $\mathcal{M}^r$ is active}
       \State Find $a_m$ by solving~\eqref{eq.action_correction} \Comment{Action correction}
    \Else 
        \State $a_m = a_t$
    \EndIf
    \State Add data sample $D = (s_t,a_m,s'_t,r,c)$ to $\mathcal{D}$ and $\mathcal{D}^e$
    \State $j = j+1$
\EndWhile
\State Construct the ROMDP $\mathcal{M}^r$ based on $\mathcal{D}$ and $\mathcal{D}^e$
\State Update the policy $\pi_\theta$ with the selected SRL algorithm
\If{loss value of $V_C(s) < \delta_d$ and $\mathcal{M}^r$ is active}
\State Deactivate $\mathcal{M}^r$ \Comment{Dynamic activation mechanism}
\ElsIf{loss value of $V_C(s) > \delta_r$ and $\mathcal{M}^r$ is not active}
\State Reactivate $\mathcal{M}^r$
\EndIf
\State $\mathcal{D}^e = \varnothing$, $i = i+1$
\EndWhile
\end{algorithmic}
\end{algorithm}

\subsection{Dynamic Activation of GenSafe}

As the dataset $\mathcal{D}$ enlarges, the prediction accuracy of the ROMDP is expected to improve. 
However, due to the inherent information loss in the order reduction process, discrepancies between the estimated and actual costs could persist regardless of the data volume. 
In contrast, with a sufficient amount of data, traditional methods of approximating value functions $V_R(s)$ and $V_C(s)$ via neural networks are able to provide accurate estimations, thereby leading to satisfactory performance of SRL algorithms.
Thus, while the utilization of GenSafe is able to enhance the safety performance of SRL algorithms under limited data availability, direct approximation of the original cost function using a neural network is expected to yield better performance in later learning phases when more data becomes available.

Recognizing this characteristic, we therefore introduce a dynamic activation mechanism for GenSafe (see Algorithm~\ref{alg.srl_gensafe}). 
The rationale behind this mechanism is that, when the direct approximation of the cost value function $V_C(s)$ demonstrates higher confidence in accuracy in later learning phases, we disable the ROMDP-based action correction to mitigate the influence of information loss caused by order reduction and to accelerate the learning process.
To achieve this, we assess the accuracy of the current cost value function $V_C(s)$ approximation after each epoch update. 
This can be done by determining whether the loss value from updating the neural network falls below a predetermined threshold. 
If it does, we deactivate GenSafe, meaning no action corrections will be applied, and the action chosen by the SRL policy $\pi_\theta$ is directly utilized. 
Subsequently, we continue to monitor the loss value associated with approximating the cost value function $V_C(s)$ throughout the learning process. 
If this value again exceeds a certain threshold, indicating a degradation in approximation accuracy, GenSafe is reactivated.
Such a dynamic activation mechanism enhances computational efficiency and ensures that the agent's performance is not overly affected by the order reduction process. 
However, it should be noted that continuous switching between the activation status of GenSafe may result in instability during policy updates. 
To mitigate this issue, a practical solution is to increase the reactivation threshold and maintain a redundancy buffer between the deactivation and reactivation thresholds. 
This allows for a reasonable degree of performance oscillation over several iterations when disabling GenSafe, thereby reducing the risk of performance instability.

\begin{remark}
Different criteria, e.g., the number of training epochs completed or the observed average cost, can also be employed to determine the deactivation or reactivation of GenSafe. 
Choosing these criteria should consider the specific requirements and properties of the task, alongside the characteristics of the selected SRL algorithm, to ensure a satisfying performance of GenSafe tailored to the learning context.
\end{remark}

\begin{figure*}[!t]
\centering
\captionsetup[subfloat]{labelfont=scriptsize,textfont=scriptsize}
\subfloat[Circle Level 1]{\includegraphics[width=.22\linewidth, height=1.1in]{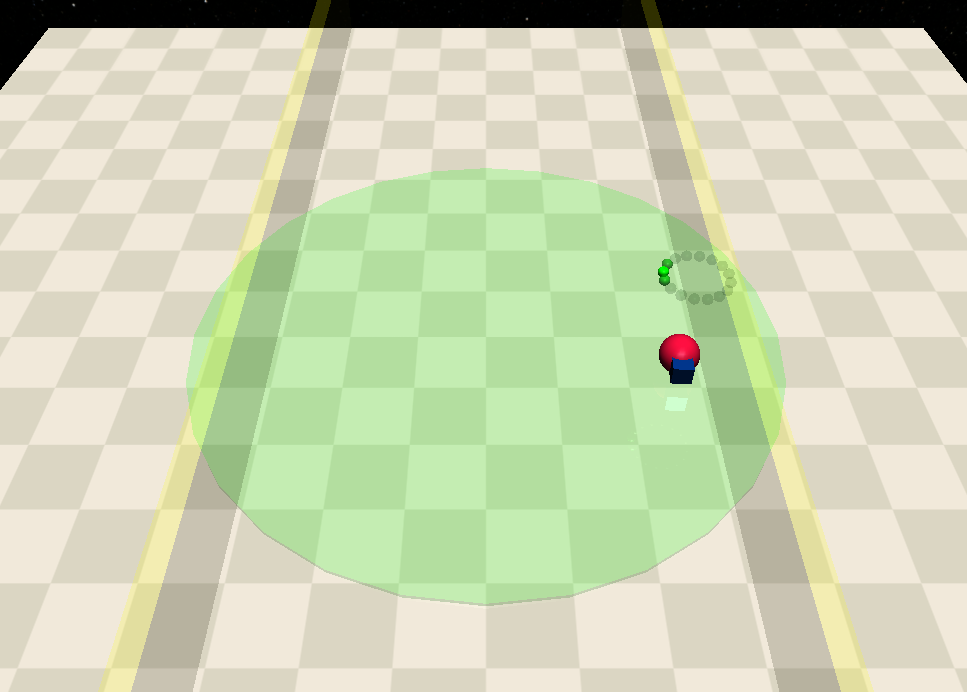}%
\label{fig.circle1}}
\hfil
\subfloat[Circle Level 2]{\includegraphics[width=.22\linewidth, height=1.1in]{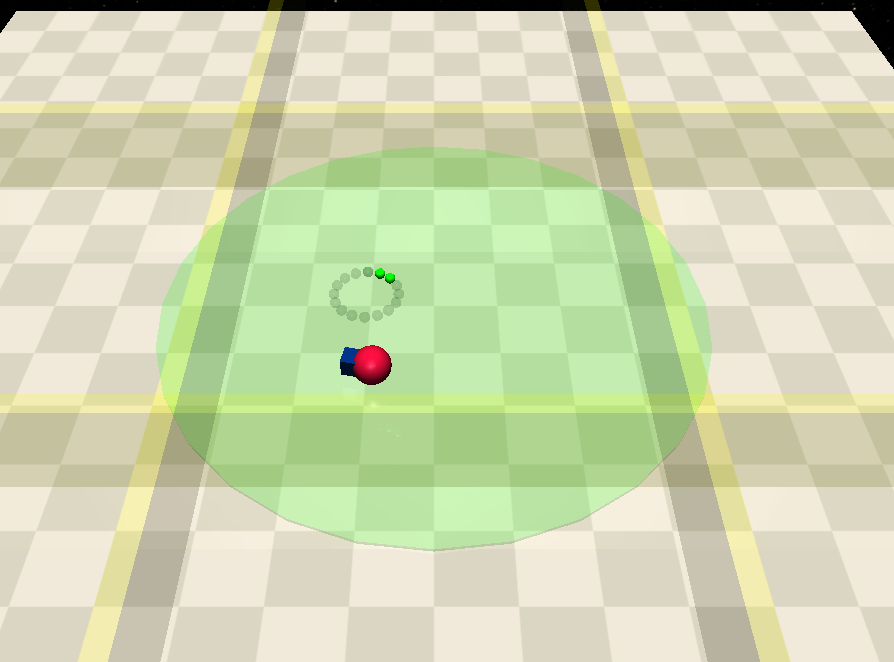}%
\label{fig.circle2}}
\hfil
\subfloat[Button Level 1]{\includegraphics[width=.22\linewidth, height=1.1in]{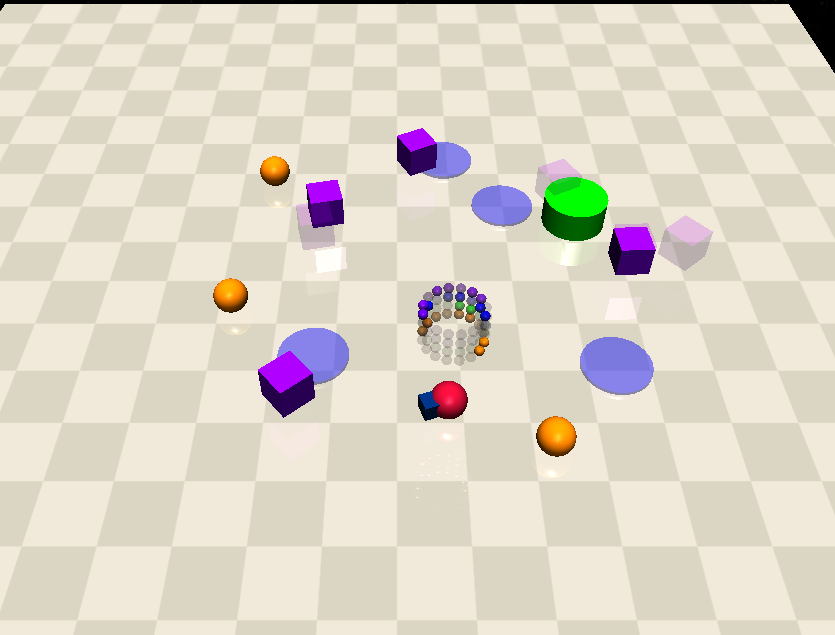}%
\label{fig.button1}}
\hfil
\subfloat[Button Level 2]{\includegraphics[width=.22\linewidth, height=1.1in]{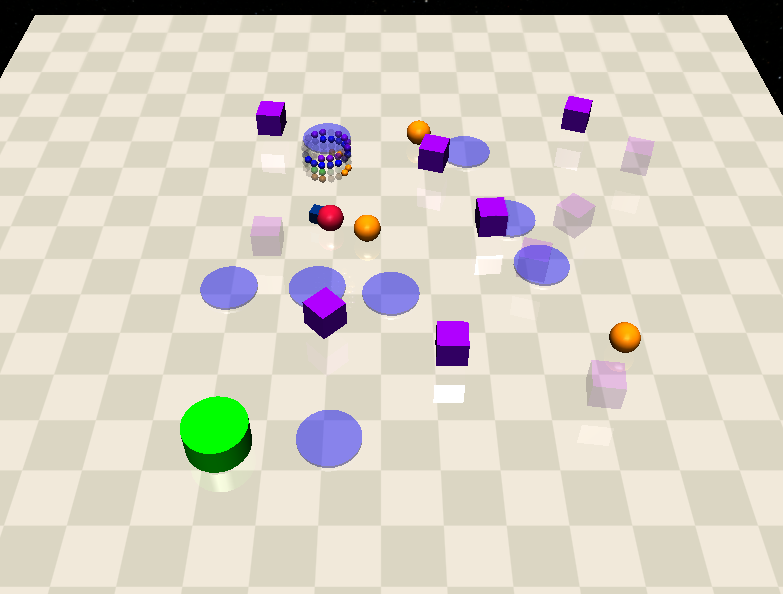}%
\label{fig.button2}}
\hfil
\subfloat[Goal]{\includegraphics[width=.22\linewidth, height=1.1in]{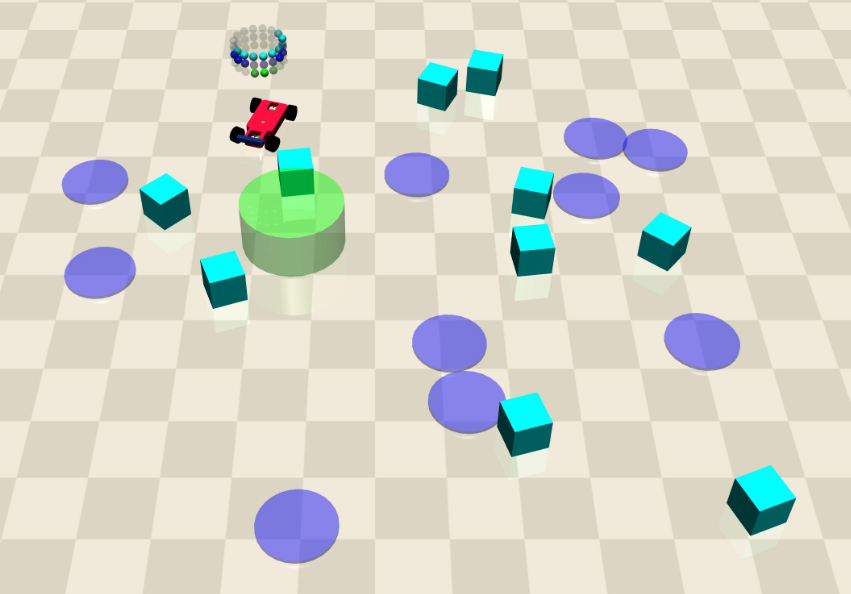}%
\label{fig.goal2}}
\hfil
\subfloat[Push]{\includegraphics[width=.22\linewidth, height=1.1in]{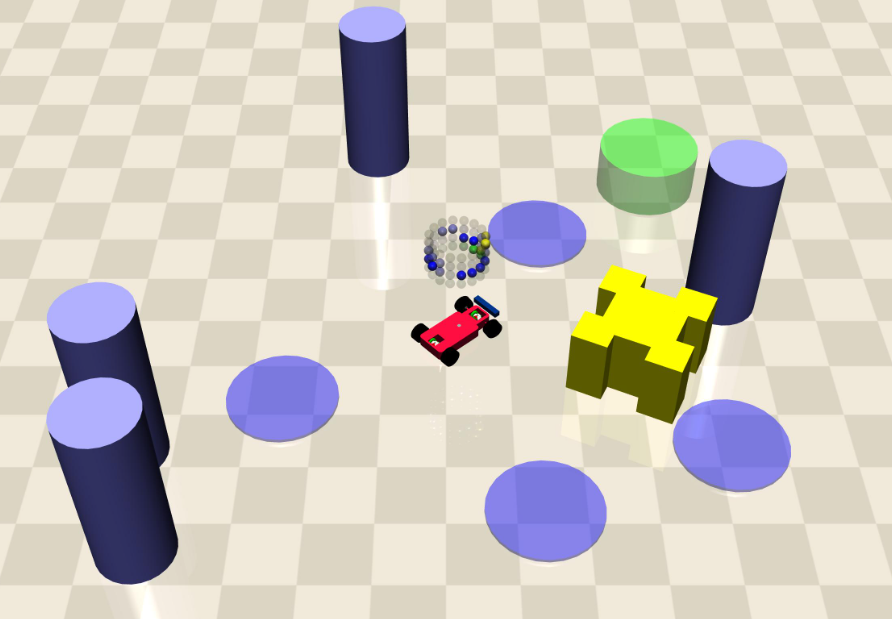}%
\label{fig.push2}}
\hfil
\subfloat[Swimmer]{\includegraphics[width=.22\linewidth, height=1.1in]{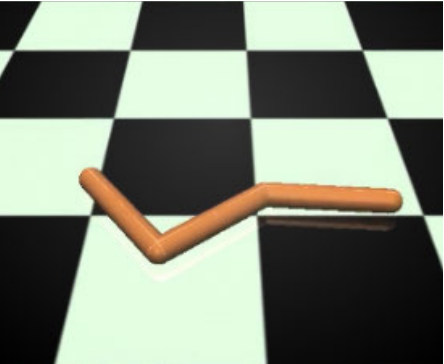}%
\label{fig.swimmer}}
\hfil
\subfloat[Hopper]{\includegraphics[width=.22\linewidth, height=1.1in]{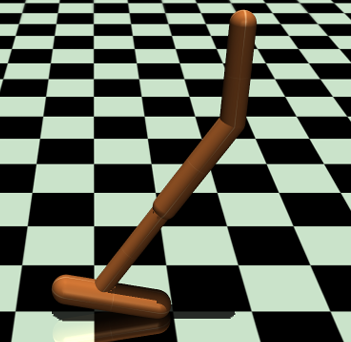}%
\label{fig.hopper}}
\caption{Eight SRL tasks used in our experiments. Readers are referred to the supplementary material and~\cite{ji2023safety} for more details about these task environments.}
\label{fig.SRL_tasks}
\vspace{-10pt}
\end{figure*}

\subsection{Practical Implementation Details}

During the practical implementation of GenSafe, we adopt several strategies to improve computational efficiency. 
Firstly, instead of reconstructing the ROMDP at every epoch, we establish a periodic update interval, meaning that the ROMDP is reconstructed only after a specified number of epochs.  
Secondly, considering the available computational resources and memory constraints, we impose a limit on the maximum number of data samples that can be stored in the dataset $\mathcal{D}$. 
Upon reaching this limit, we prioritize the retention of newer data samples and discard older entries to maintain the dataset size.
Thirdly, given that the computational cost of t-SNE increases with more data samples, we also limit the maximum number of data samples employed in t-SNE computations. 
If the number of data samples in $\mathcal{D}$ exceeds this limit, we randomly select a subset from $\mathcal{D}$ that matches this limit for t-SNE analysis, thereby maintaining manageable computational loads. 
The data not used remains part of $\mathcal{D}$ and is utilized in other abstraction processes, ensuring that only the t-SNE computations are affected by this subsampling.
These measures collectively enhance the computational efficiency of GenSafe. 
However, the determination of the parameter values should take into account the available computational resources as well as the specific requirements of the tasks at hand.

\section{Experimental Results}
\label{sec.results}

In this section, we evaluate the performance of the proposed GenSafe on various SRL benchmark problems. 
More details are provided as follows.

\subsection{Experimental Setup}

We assess the performance of GenSafe on eight representative SRL tasks that are given in Safety-Gymnasium~\cite{ji2023safety}, a collection of SRL environments that builds upon the widely recognized Safety-Gym~\cite{ray2019benchmarking} benchmark. 
The tasks we used are: \textit{Circle Level 1}, \textit{Circle Level 2}, \textit{Button Level 1}, \textit{Button Level 2}, \textit{Goal}, \textit{Push}, \textit{Swimmer}, and \textit{Hopper} (see Fig.~\ref{fig.SRL_tasks}).
Due to page limit, readers are referred to the supplementary material and~\cite{ji2023safety} for more details about these task environments.

For each task, we compare the performance of eleven different learning methods: (1) \textit{PPO}, (2) \textit{PPO-Lagrangian (PL)}, (3) \textit{PPO-Lagrangian with GenSafe (PL-G)}, (4) \textit{Safe Exploration (SE)}~\cite{dalal2018safe}, (5) \textit{PPO-Lagrangian with Neural Network Estimator (PL-NN)}, (6) \textit{FOCOPS (FO)}~\cite{zhang2020first}, (7) \textit{FOCOPS with GenSafe (FO-G)}, (8) \textit{CUP}~\cite{cup}, (9) \textit{CUP with GenSafe (CUP-G)}, (10) \textit{WCSAC (WS)}~\cite{wcsac}, and (11) \textit{WCSAC with GenSafe (WS-G)}. 
PPO serves as the reference method to illustrate the standard performance of DRL algorithms.
PL, SE, FO, CUP, and WS are well-established SRL methods ranging from earlier to more recent developments, and are considered baseline methods for evaluating SRL performance.
The SE algorithm also introduces a safety layer; however, the original version relies on offline-collected data. 
To make a fair comparison, we modify SE to generate the safety layer using online-collected data instead.
We assess the effectiveness of GenSafe by comparing the performance of the respective GenSafe-augmented algorithms (PL-G, FO-G, CUP-G, and WS-G) against their original SRL counterparts (PL, FO, CUP, and WS). 
Additionally, we implement PL-NN by replacing the ROMDP in PL-G with a simple neural network cost estimator. 
The comparison between PL-NN and PL-G serves as an ablation study, demonstrating the ROMDP's effectiveness in producing accurate safety estimates with limited data, thereby validating the performance of our proposed approach.
To ensure a thorough comparison, each method is executed on each task using five randomly selected seeds, and the results are averaged accordingly.
Details of the parameters used in our experiments are provided in the supplementary material.

\subsection{Model Order Reduction}

\begin{figure}[!t]
\centering
\captionsetup[subfloat]{labelfont=scriptsize,textfont=scriptsize}
\subfloat[]{\includegraphics[trim={7mm 0mm 7mm 3mm}, width=.48\linewidth]{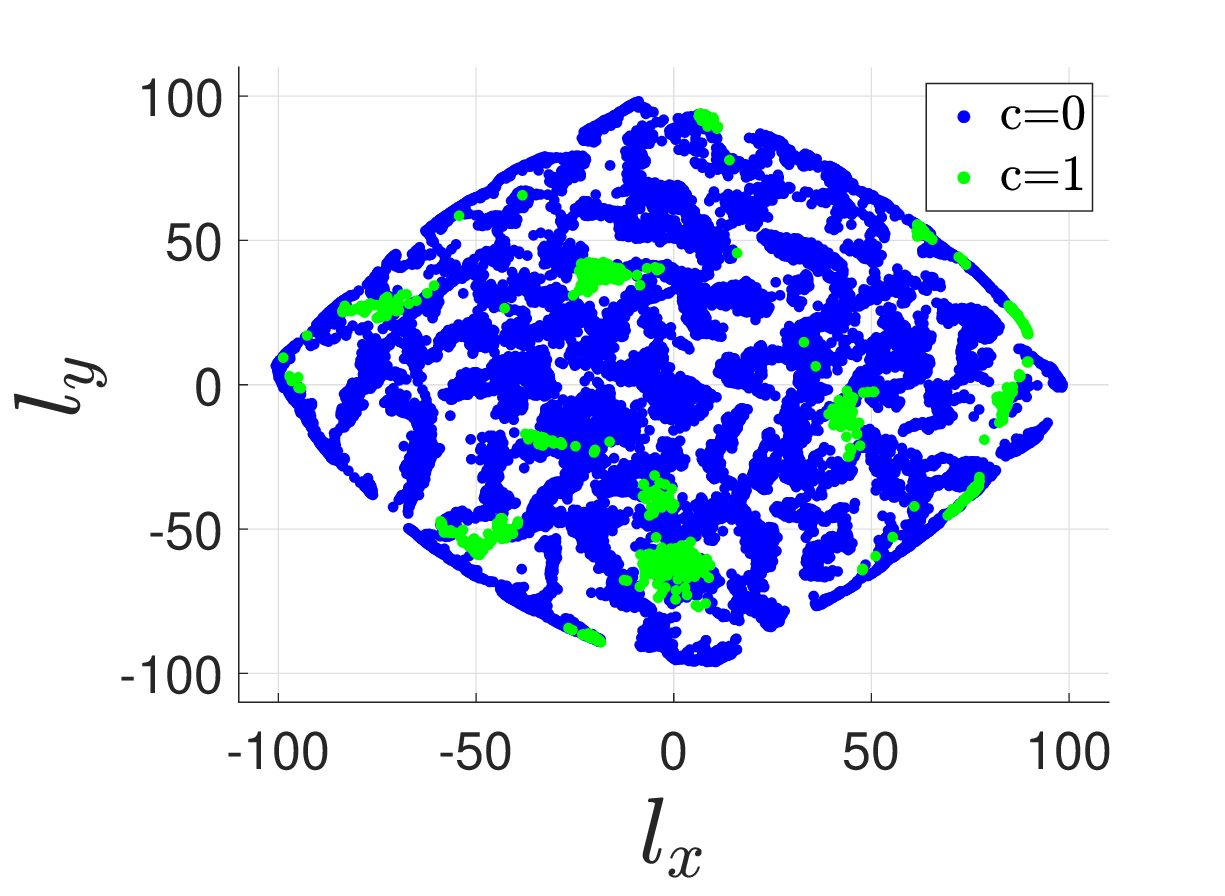}%
\label{fig.low_states}}
\hfil
\subfloat[]{\includegraphics[trim={7mm 0mm 7mm 10mm}, width=.48\linewidth]{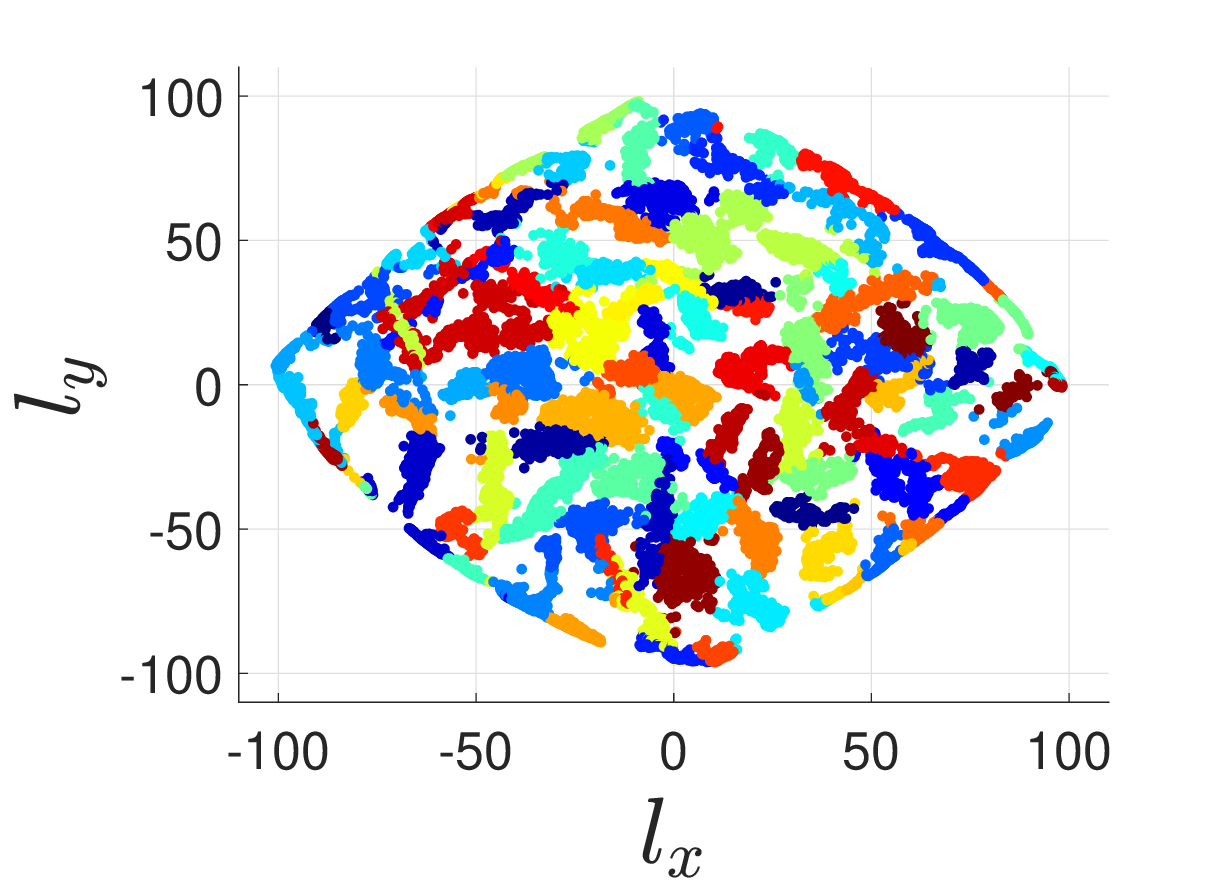}%
\label{fig.low_state_index}}
\caption{(a) The set of low-dimensional states $\mathcal{D}_{l} = \{l_1, \ldots, l_{20000}\}$ distributed among a two-dimensional state space $S^l$, which is derived by applying t-SNE on the set of original states $\mathcal{D}_{s} = \{s_1, \ldots, s_{20000} \}$. The dimensions $l_x$ and $l_y$ represent two abstract features of these low-dimensional states. Each low-dimensional state corresponds to an observed high-dimensional data that has a cost of either $c=0$ (blue) or $c=1$ (green), where $c=0$ and $c=1$ denote safe and unsafe observations, respectively. 
(b) Results of the GMM classifier. Each low-dimensional state $l_i$ is categorized into one of the $k_s=100$ cluster regions, with each region represented by a different color and assigned an index $v_s \in \{1,2,\ldots,100 \}$. The reduced state $s^r$ corresponding to each original state $s$ is therefore given by the cluster index, i.e., $s^r = v_s = f_s(s) = f_\text{GMM}(l) = f_\text{GMM}(f_l(s))$. This results in a dataset of reduced states $\mathcal{D}_{s^r} = \{s^r_1,\ldots,s^r_{20000}\}$.  }
\label{fig.order_reduction}
\vspace{-10pt}
\end{figure}

We utilize the initial construction of ROMDP in the \textit{Button Level 1} task as an example to illustrate the results of model order reduction, i.e., the application of t-SNE for deriving a reduced state space. 
Given a dataset $\mathcal{D}= \{D_1, \ldots, D_n \}$ of $n=20000$ data samples observed during the first learning iteration, the associated high-dimensional states and costs are defined as $\mathcal{D}_s = \{s_1, \ldots, s_n \}$ and $\mathcal{D}_c = \{c_1, \ldots, c_n \}$, respectively. 
According to the setup of Safety-Gymnasium~\cite{ji2023safety}, the observed cost is a binary value, i.e., $c_i = \{1, 0\}$, where we consider $c_i=1$ and $c_i=0$ indicate unsafe and safe data, respectively.
We then employ t-SNE to map the high-dimensional states into a corresponding set of low-dimensional states $\mathcal{D}_{l} = \{l_1, \ldots, l_n\}$.
The results are depicted in Fig.~\ref{fig.low_states}, where each 76-dimensional original state $s_i$ is transformed into a two-dimensional state $l_i$. 
Note that, similar to outcomes of PCA, the dimensions of the low-dimensional states, i.e., $l_x$ and $l_y$, do not possess actual physical meanings but serve only as abstract features. 
The color of each low-dimensional state $l_i$ indicates its associated cost $c_i$ of the corresponding high-dimensional data $D_i$.
It can be observed that, t-SNE effectively reduces the original high-dimensional state space $S$ into a representative low-dimensional state space $S^l$, grouping safe and unsafe high-dimensional data samples closely. 
This supports the assumption that states with proximate reduced states are likely to incur similar costs, thereby formulating the basis for employing such a reduced state space to approximate the original cost function.

After determining the set of low-dimensional states $\mathcal{D}_{l} = \{l_1, \ldots, l_n\}$ via t-SNE, we proceed to train a GMM classifier on this dataset, choosing the number of clusters as $k_s=100$.
The corresponding outcomes of the GMM classifier are illustrated in Fig.~\ref{fig.low_state_index}, where each low-dimensional state $l_i$ is categorized into one of the $100$ cluster regions.
We assign each cluster region an index $v_s \in \{1,2,\ldots,100\}$, with different colors in the figure representing different clusters. 
Consequently, the reduced state $s^r$ for each original state $s$ is given by the corresponding cluster index, i.e., $s^r = v_s = f_s(s) = f_\text{GMM}(l) = f_\text{GMM}(f_l(s))$, leading to the set of reduced states $\mathcal{D}_{s^r} = \{s^r_1,\ldots,s^r_n\}$.
The reduced state space is, therefore, the collection of all these indices $S^r = \{1,2,\ldots,100\}$.  
Utilizing the identified $\mathcal{D}_{s^r}$, we accordingly construct the ROMDP. 
More details about the performance of GenSafe based on the constructed ROMDP are presented in the next subsection.

\subsection{Constraint Violations}

\begin{figure*}[!t]
\centering
\captionsetup[subfloat]{labelfont=scriptsize,textfont=scriptsize}
\subfloat[Circle Level 1]{\includegraphics[width=.25\linewidth]{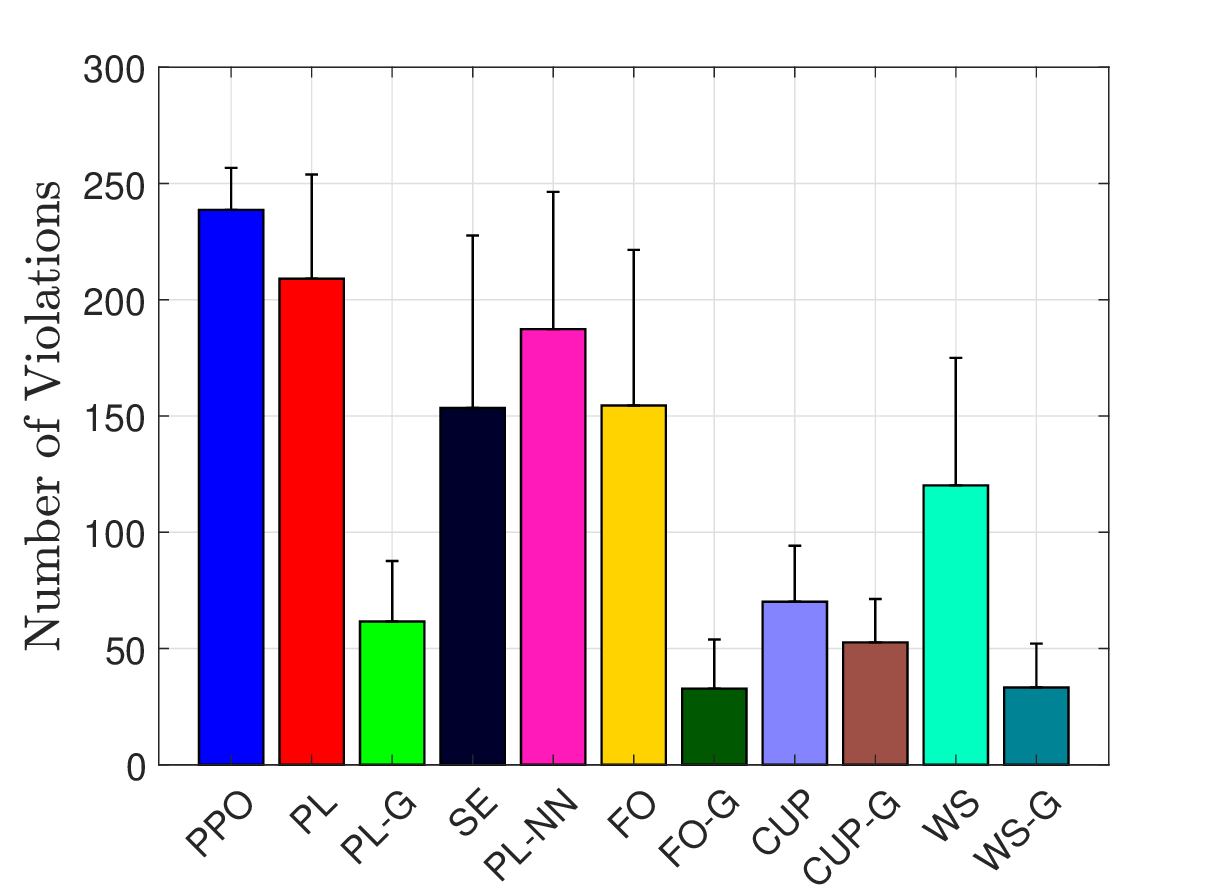}%
\label{fig.num_vio_circle1}}
\hfil
\subfloat[Circle Level 2]{\includegraphics[width=.25\linewidth]{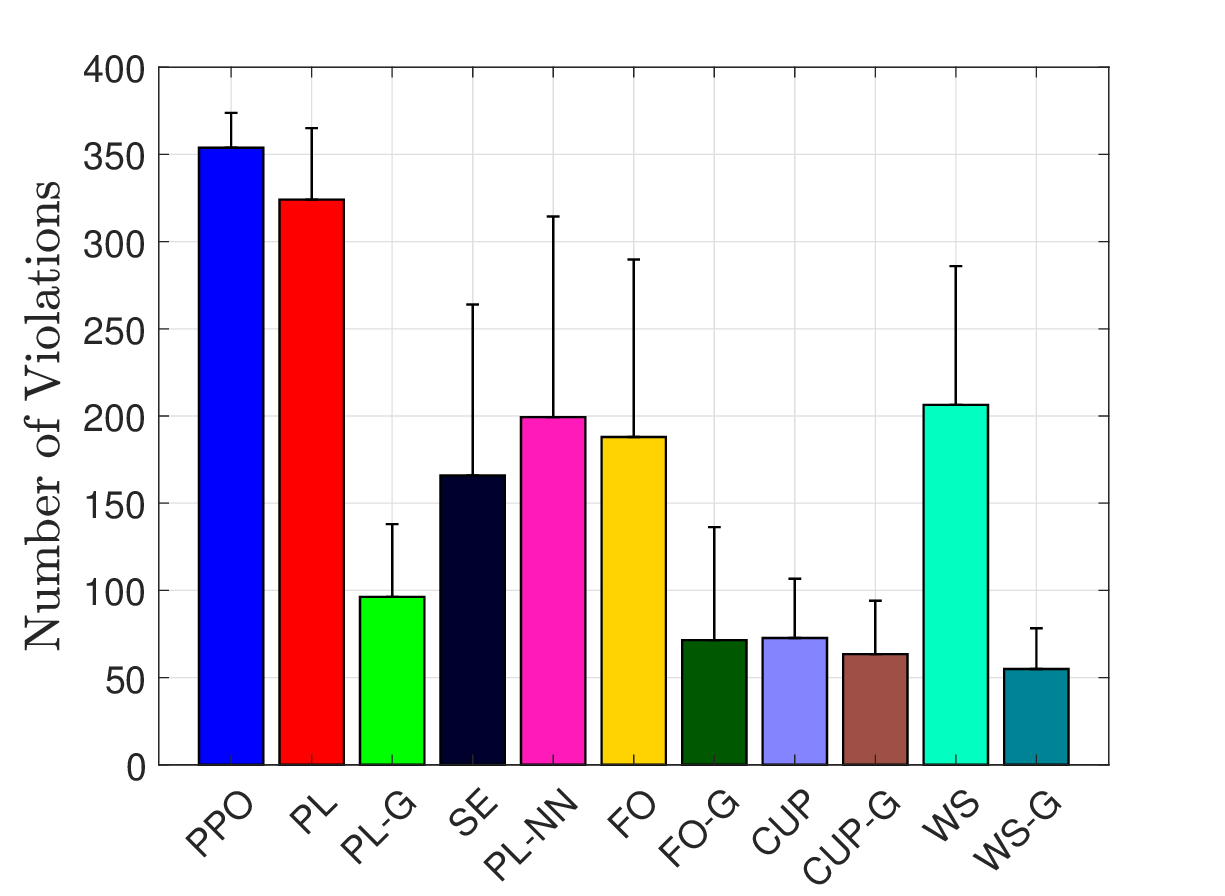}%
\label{fig.num_vio_circle2}}
\hfil
\subfloat[Button Level 1]{\includegraphics[width=.25\linewidth]{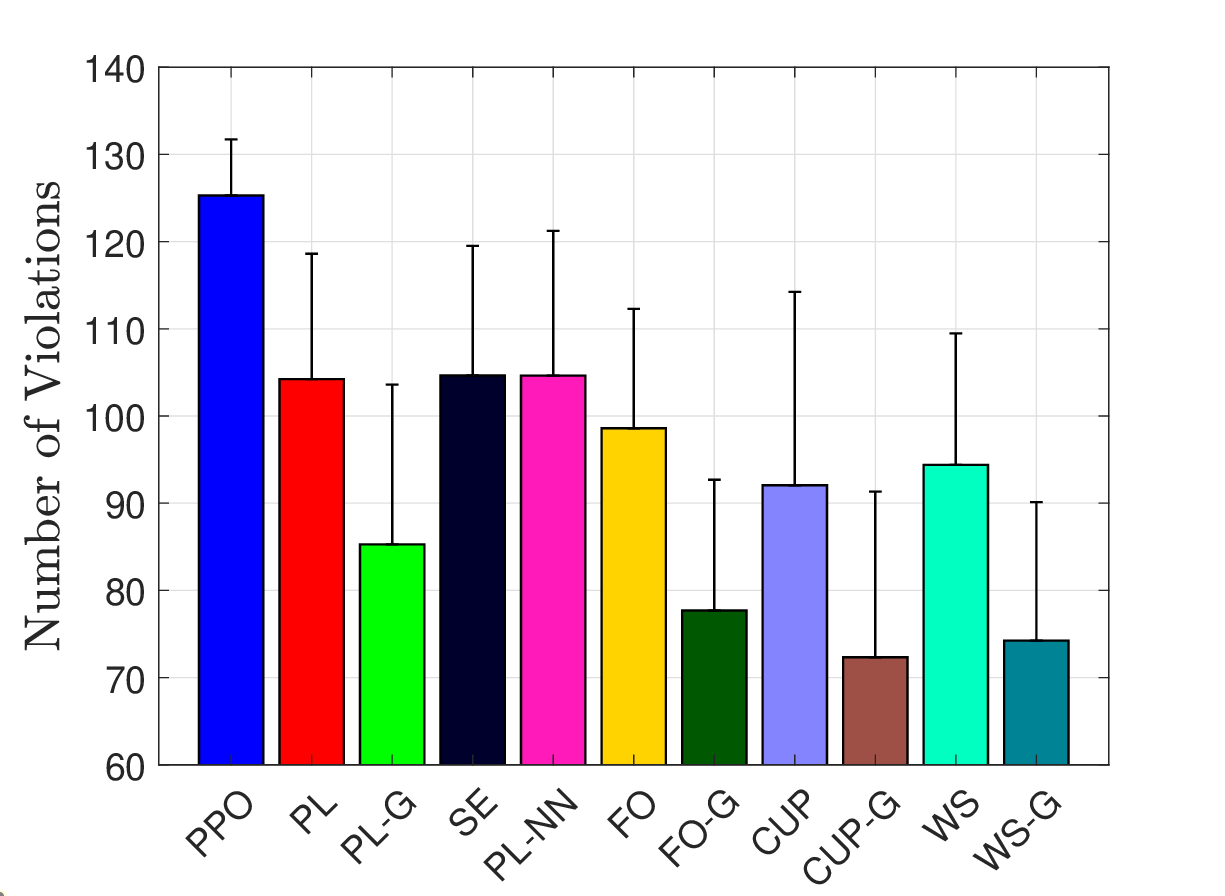}%
\label{fig.num_vio_button1}}
\hfil
\subfloat[Button Level 2]{\includegraphics[width=.25\linewidth]{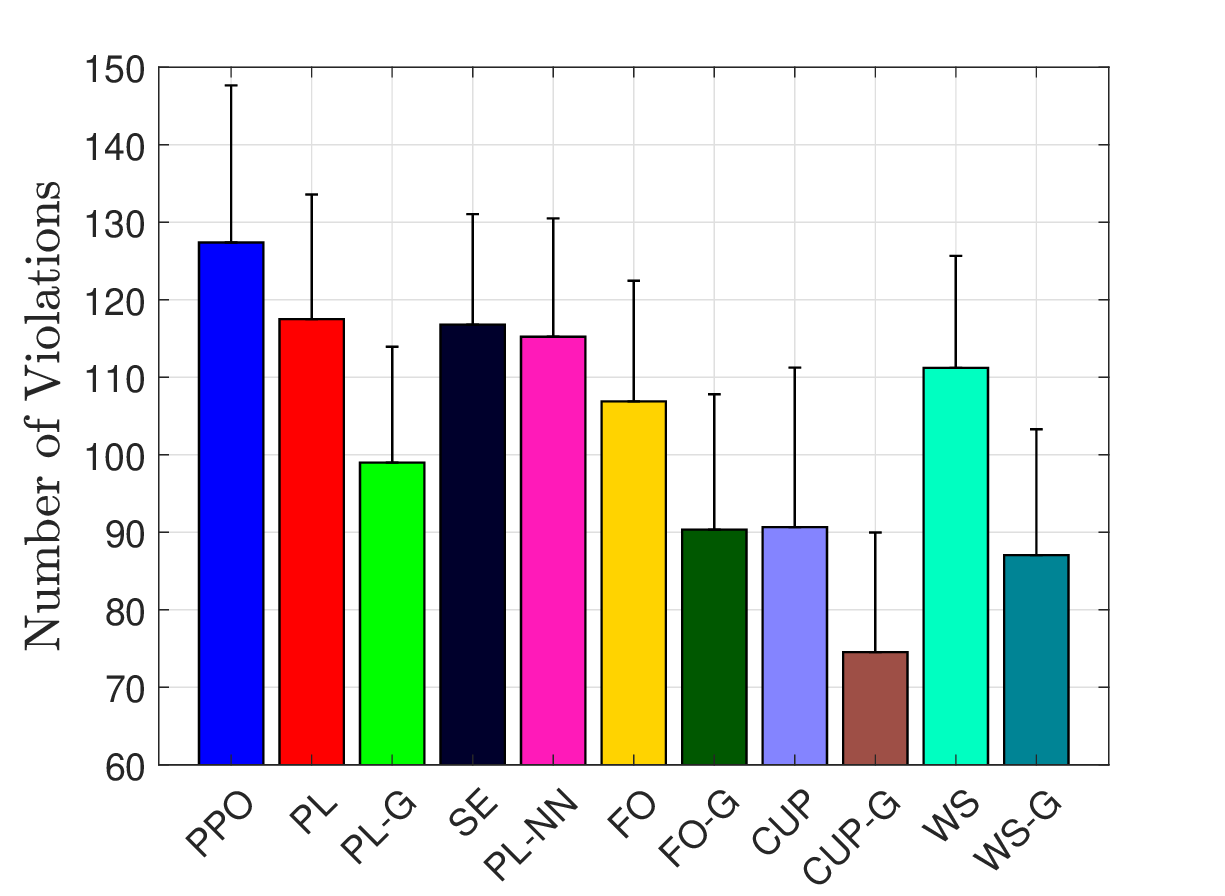}%
\label{fig.num_vio_button2}}
\hfil
\subfloat[Goal]{\includegraphics[width=.25\linewidth]{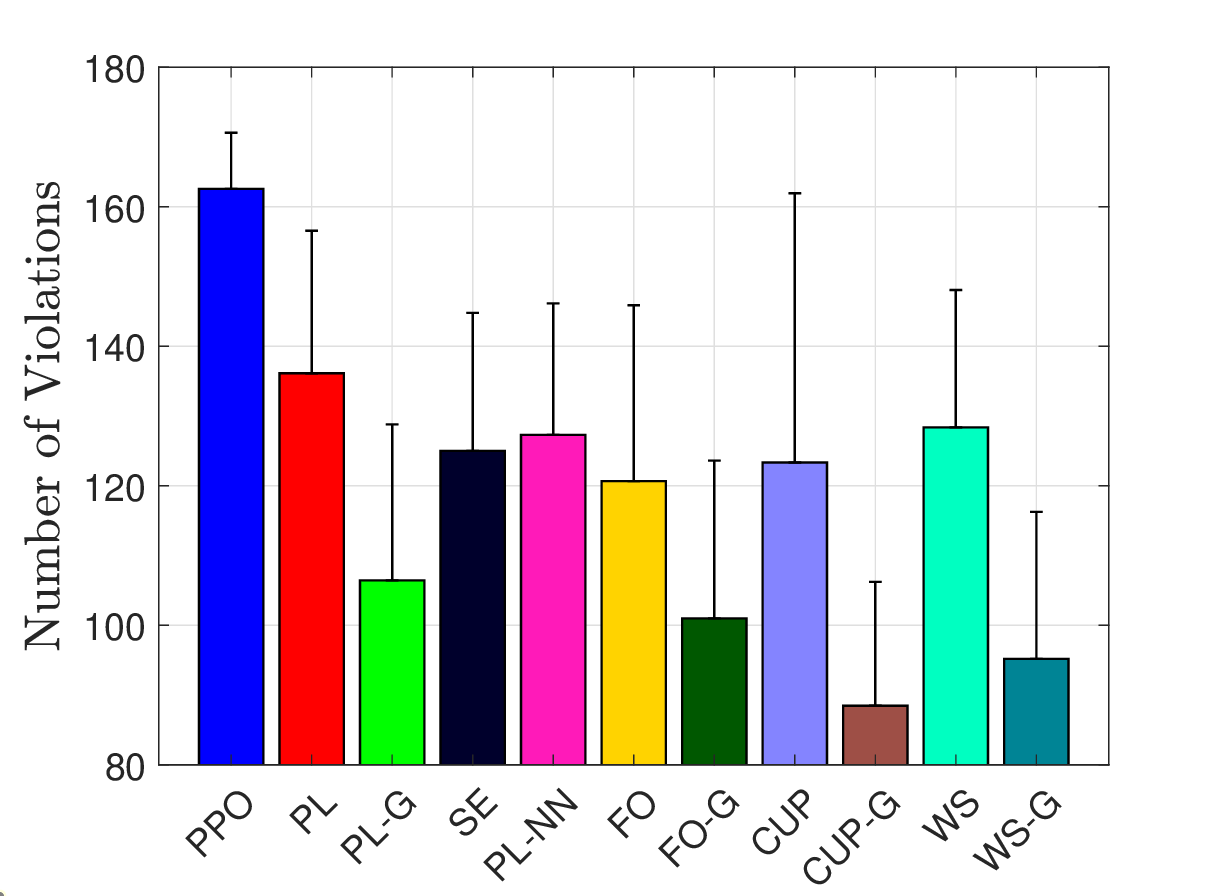}%
\label{fig.num_vio_goal2}}
\hfil
\subfloat[Push]{\includegraphics[width=.25\linewidth]{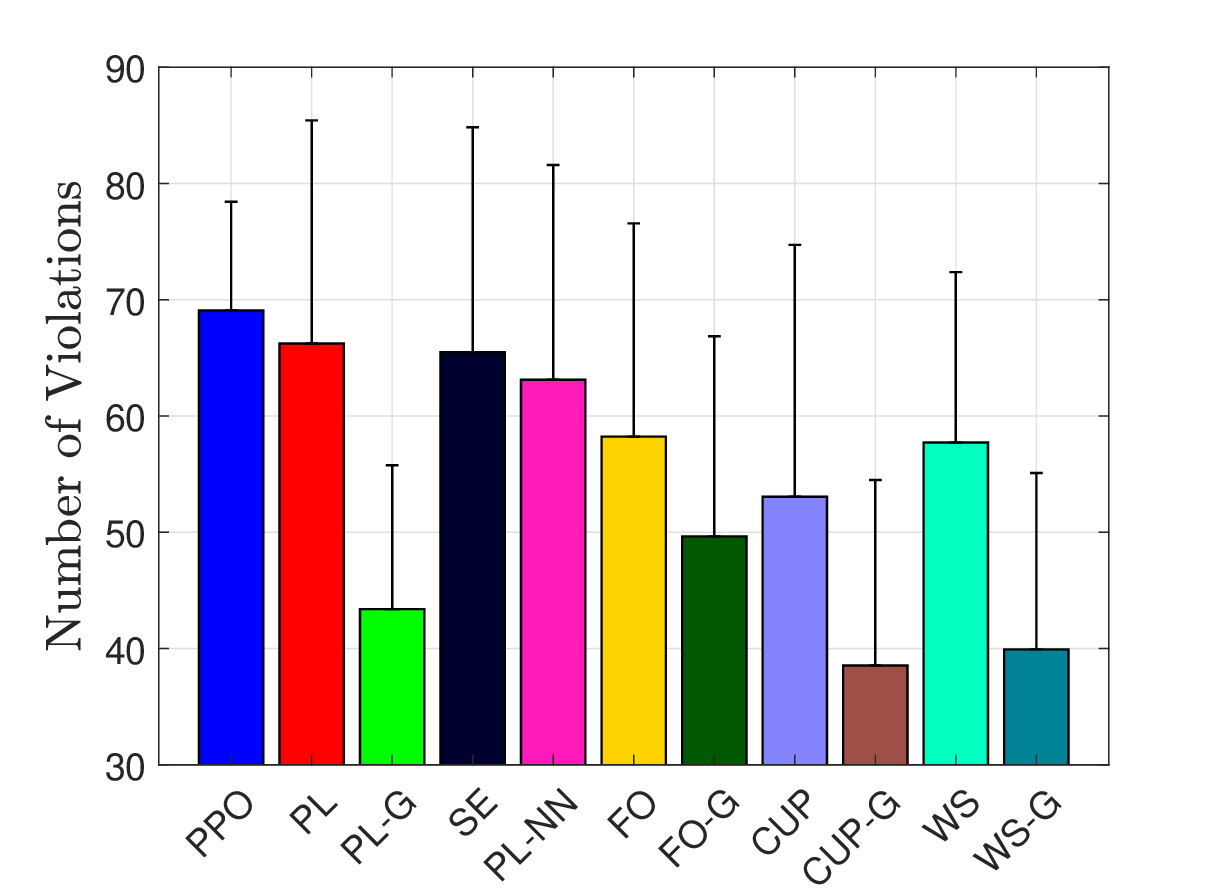}%
\label{fig.num_vio_push2}}
\hfil
\subfloat[Swimmer]{\includegraphics[width=.25\linewidth]{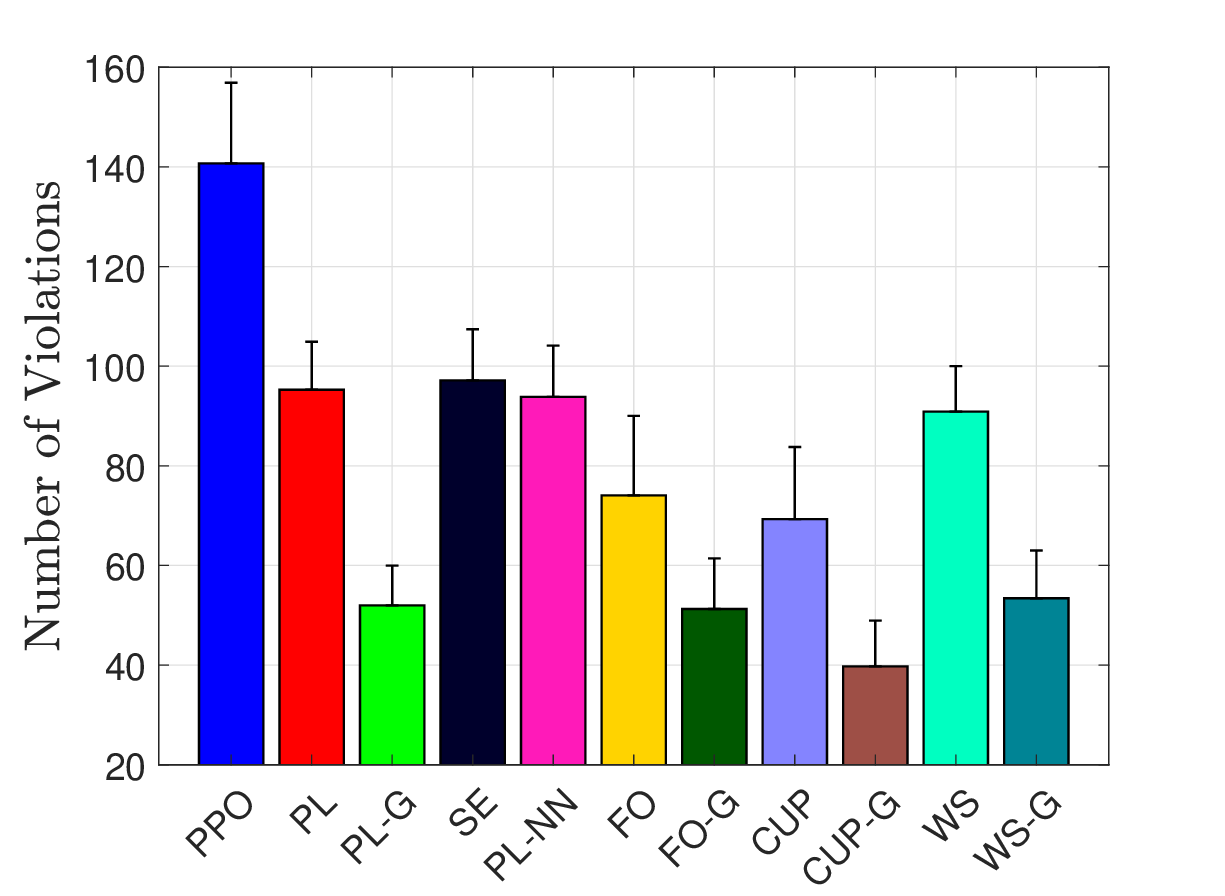}%
\label{fig.num_vio_swimmer}}
\hfil
\subfloat[Hopper]{\includegraphics[width=.25\linewidth]{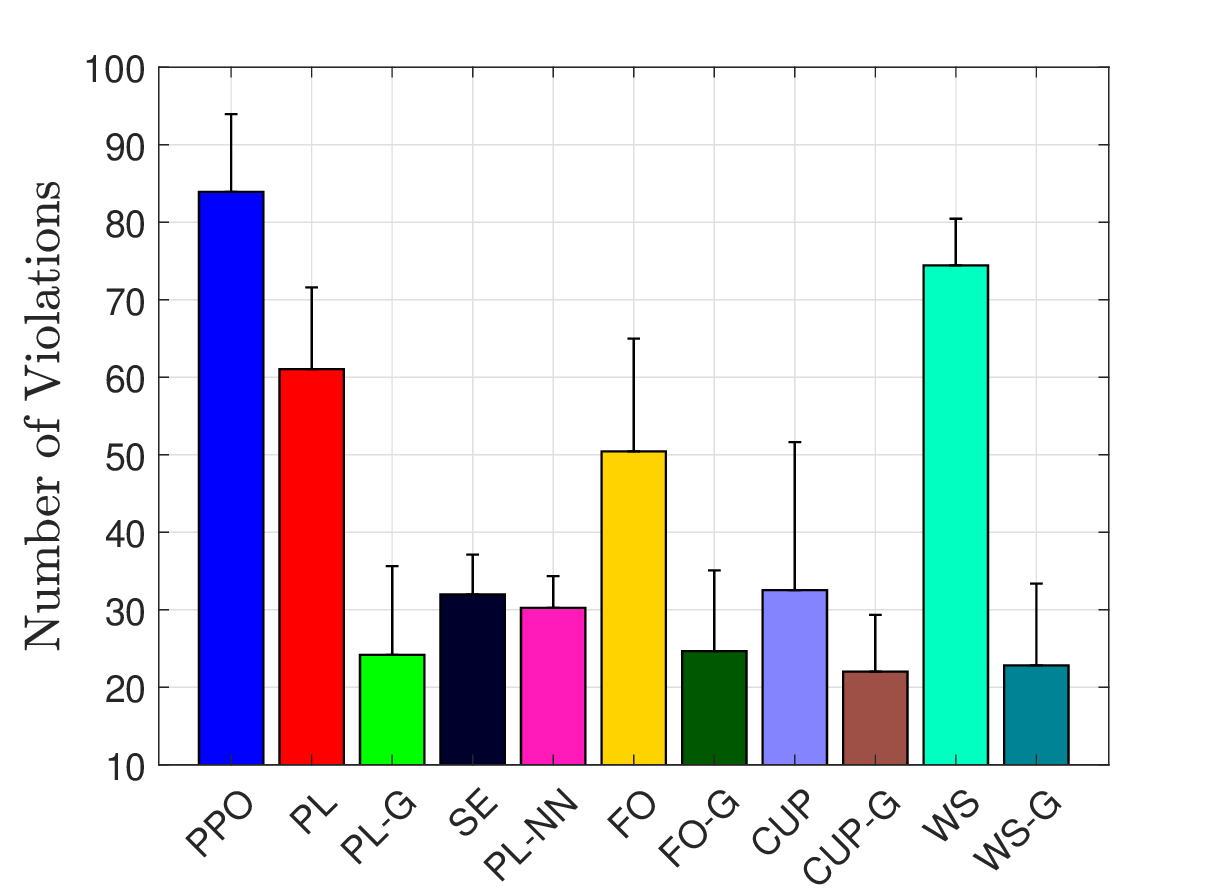}%
\label{fig.num_vio_hopper}}
\caption{The average number of safety constraint violations per epoch (20000 timesteps) over the initial 30 epochs for the eight SRL tasks.}
\label{fig.num_violations}
\vspace{-10pt}
\end{figure*}

Since the primary goal of GenSafe is to improve the safety performance of SRL algorithms during the early learning stages, we first evaluate the average number of safety constraint violations per epoch ($k_t = 20000$ timesteps) over the initial 30 epochs. 
The results are presented in Fig.~\ref{fig.num_violations}.
It can be observed that GenSafe effectively reduces the number of constraint violations of existing SRL algorithms across all eight tasks (see PL-G vs. PL, FO-G vs. FO, CUP-G vs. CUP, and WS-G vs. WS). 
While the extent of reduction varies slightly depending on the properties and difficulty of each task, GenSafe consistently demonstrates notable improvement in safety performance, highlighting its effectiveness across diverse SRL scenarios.

Due to the difficulty of generating an accurate linear cost model with limited data, the SE algorithm yields less satisfactory safety performance. 
Similarly, a simple neural network cost estimator struggles to model the complex cost functions for the tested SRL tasks, resulting in minimal improvements in safety. 
In contrast, GenSafe utilizes order reduction techniques and ROMDP to overcome data insufficiency, leading to more reliable cost estimates. 
The remarkable reduction in constraint violations observed between SE, PL-NN, and PL-G during the early learning phases demonstrates the effectiveness of our proposed ROMDP and GenSafe approaches.

\subsection{Task and Safety Performance}

\begin{figure*}[!t]
\centering
\captionsetup[subfloat]{labelfont=scriptsize,textfont=scriptsize}
\subfloat[Circle Level 1]{\begin{tabular}[b]{c}
\includegraphics[trim={7mm 0mm 7mm 3mm}, width=.22\linewidth]{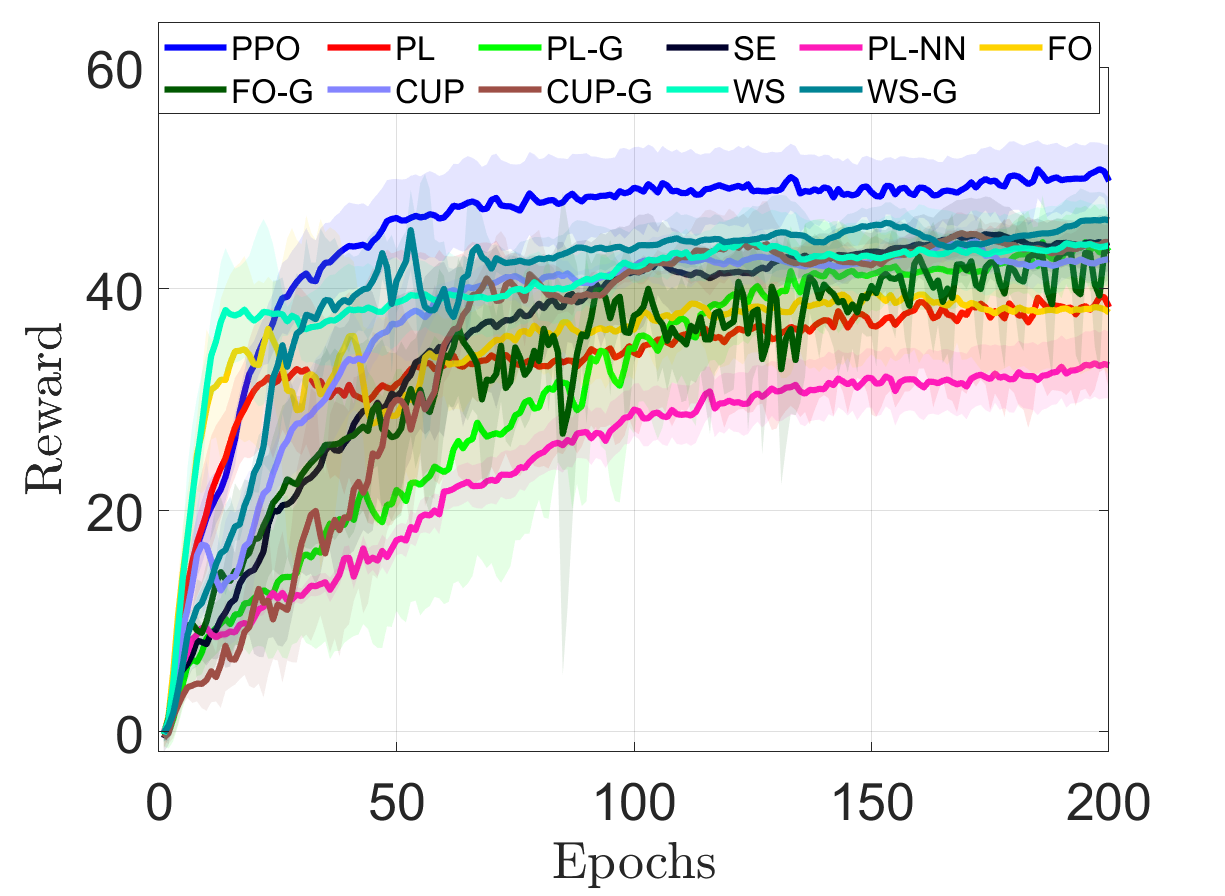}\\[-9pt]
\includegraphics[trim={7mm 0mm 7mm 3mm}, width=.22\linewidth]{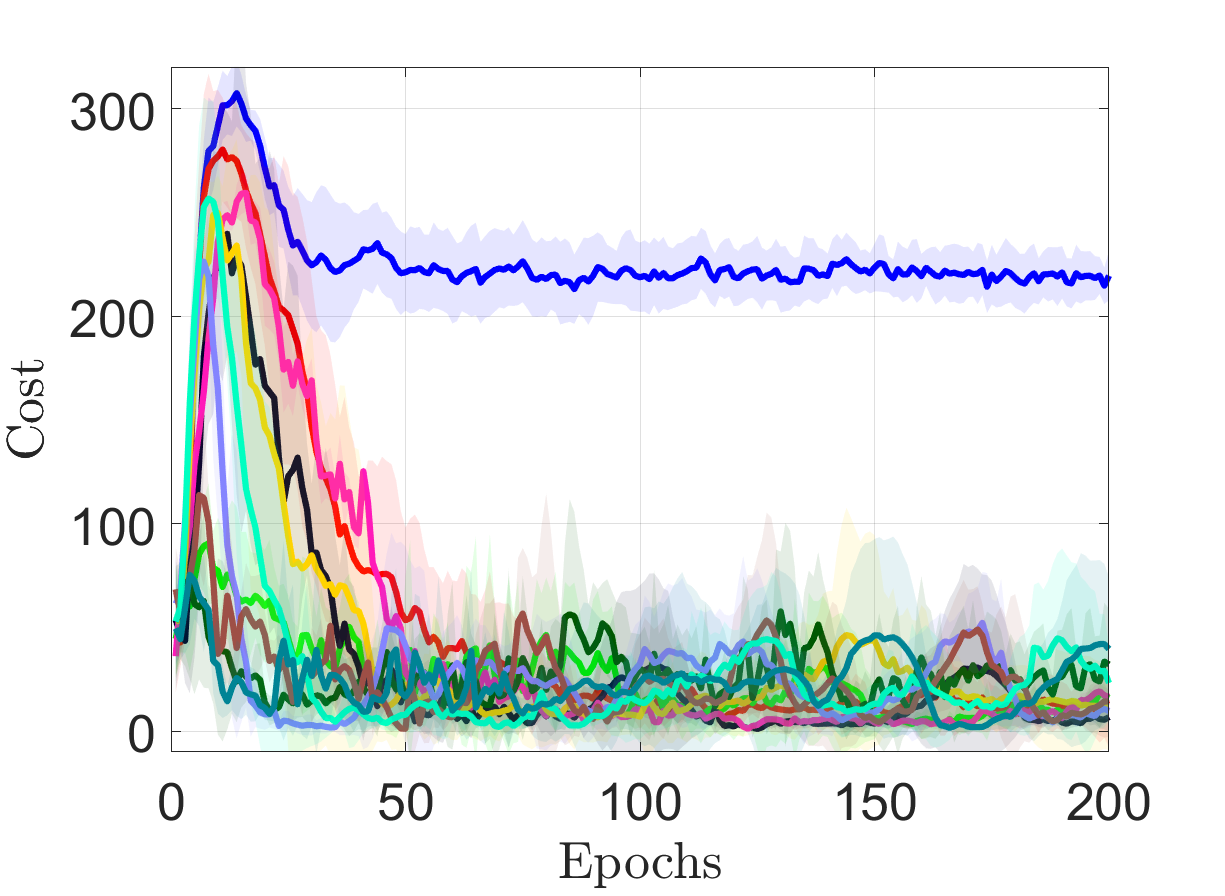}
\end{tabular}%
\label{fig.results_circle1}}
\hfil
\subfloat[Circle Level 2]{\begin{tabular}[b]{c}
\includegraphics[trim={7mm 0mm 7mm 3mm}, width=.22\linewidth]{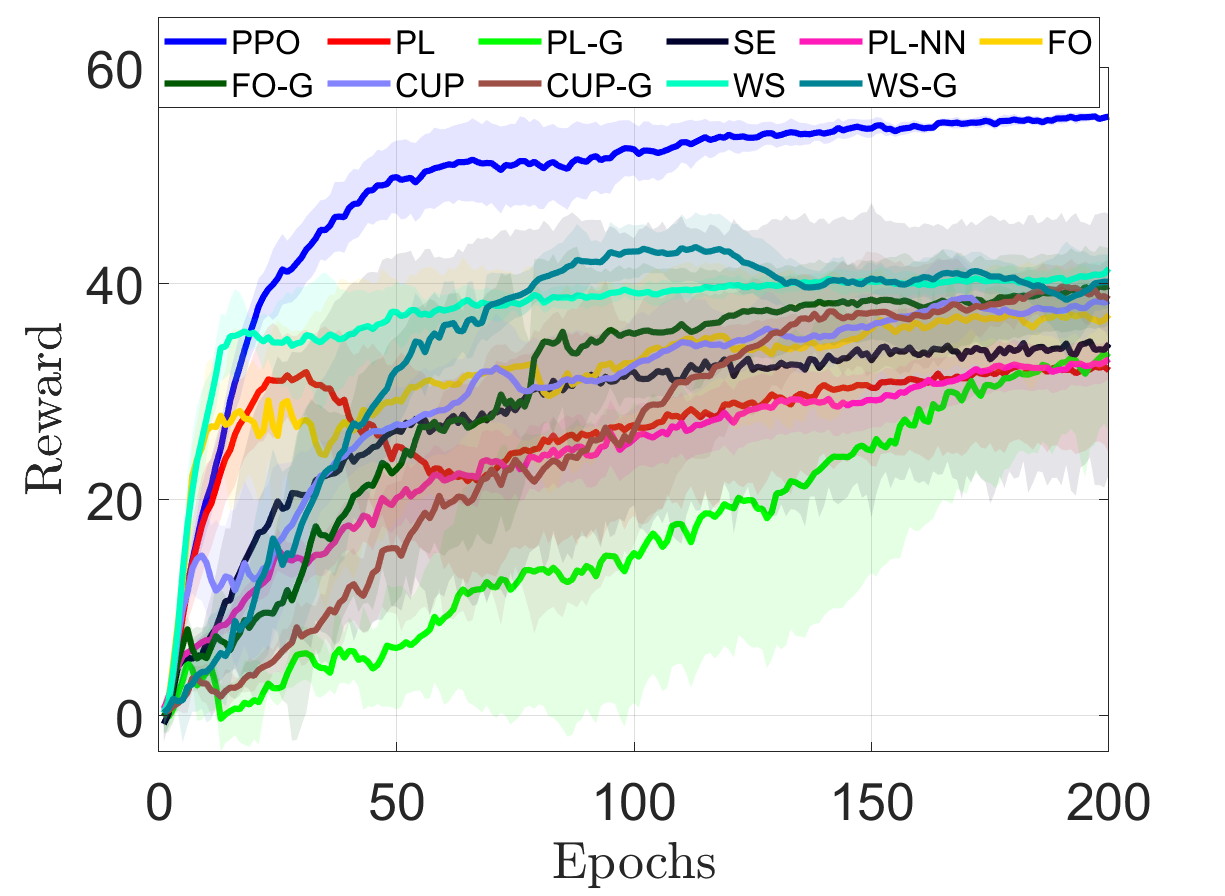}\\[-9pt]
\includegraphics[trim={7mm 0mm 7mm 3mm}, width=.22\linewidth]{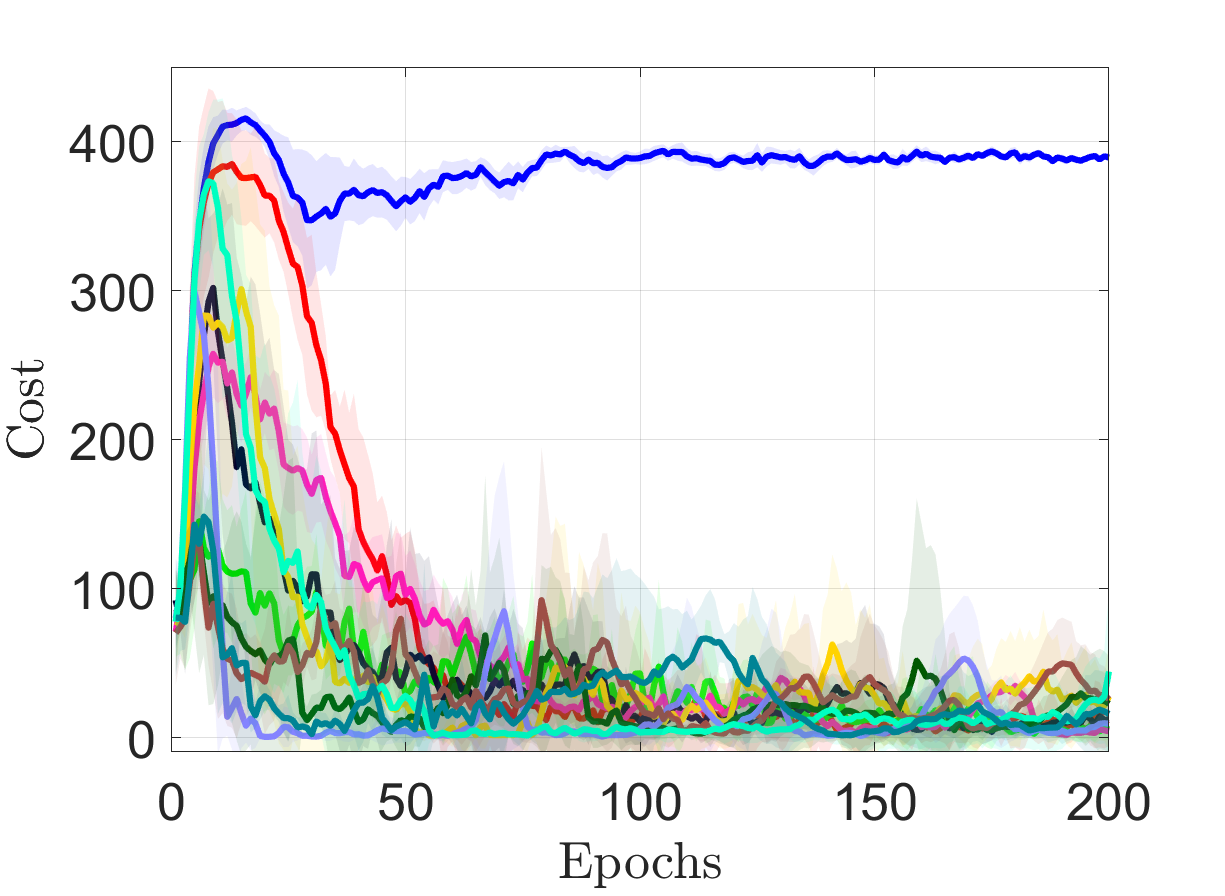}
\end{tabular}%
\label{fig.results_circle2}}
\hfil
\subfloat[Button Level 1]{\begin{tabular}[b]{c}
\includegraphics[trim={7mm 0mm 7mm 3mm}, width=.22\linewidth]{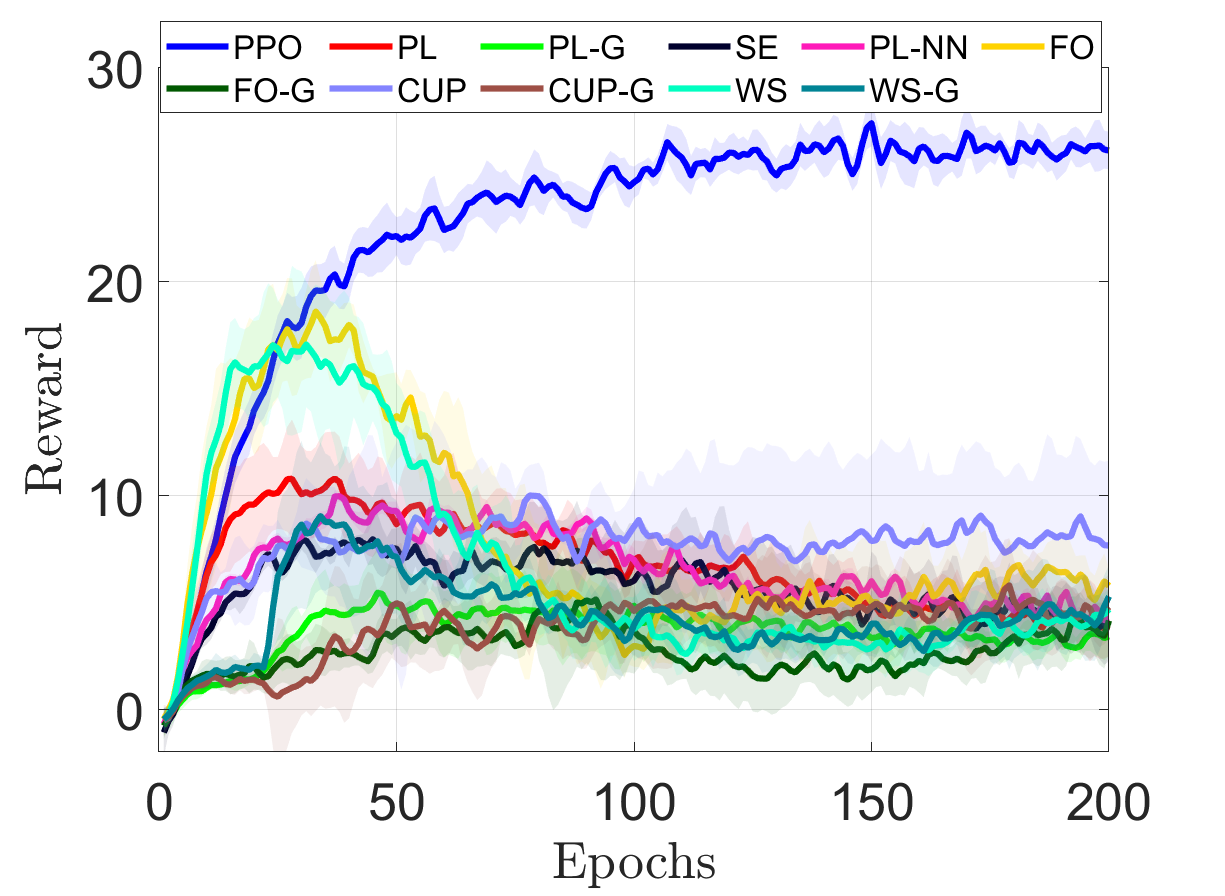}\\[-9pt]
\includegraphics[trim={7mm 0mm 7mm 3mm}, width=.22\linewidth]{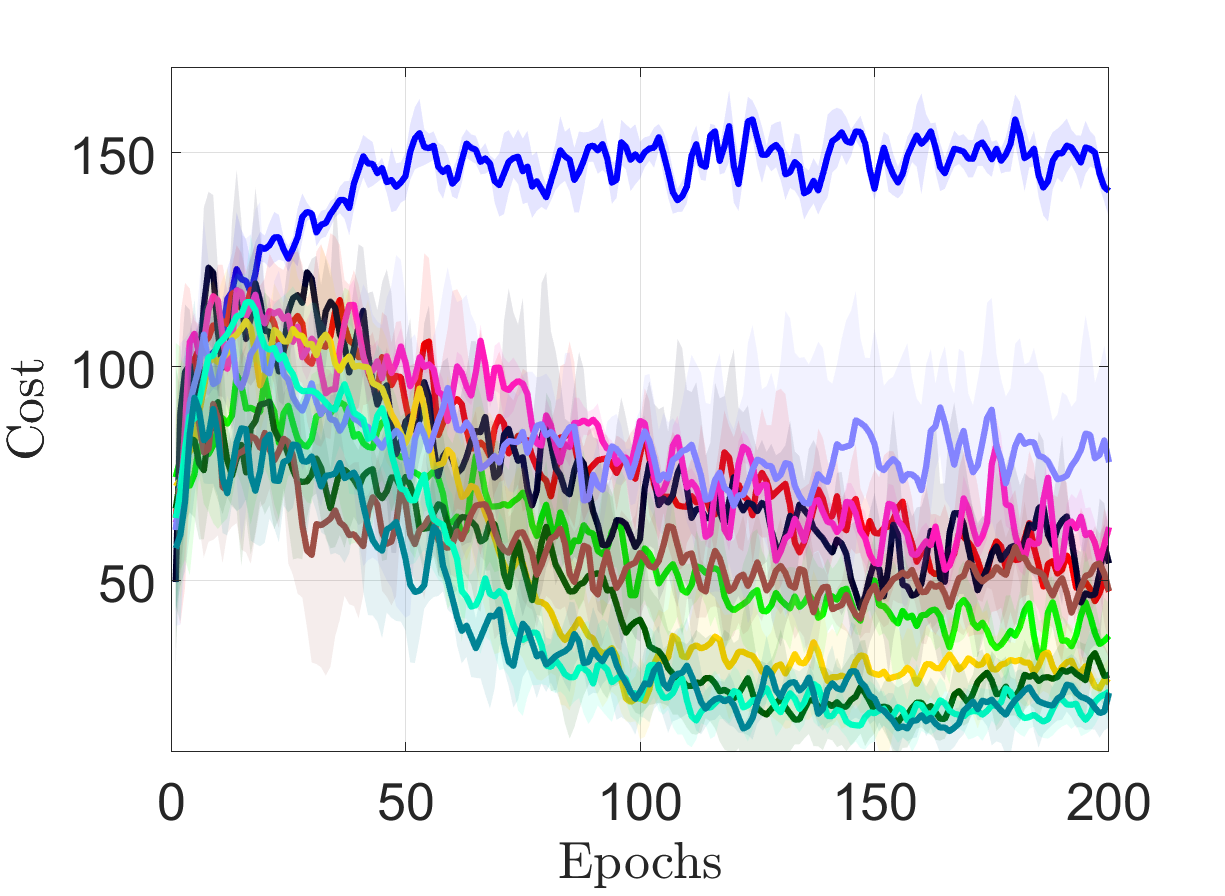}
\end{tabular}%
\label{fig.results_button1}}
\hfil
\subfloat[Button Level 2]{\begin{tabular}[b]{c}
\includegraphics[trim={7mm 0mm 7mm 3mm}, width=.22\linewidth]{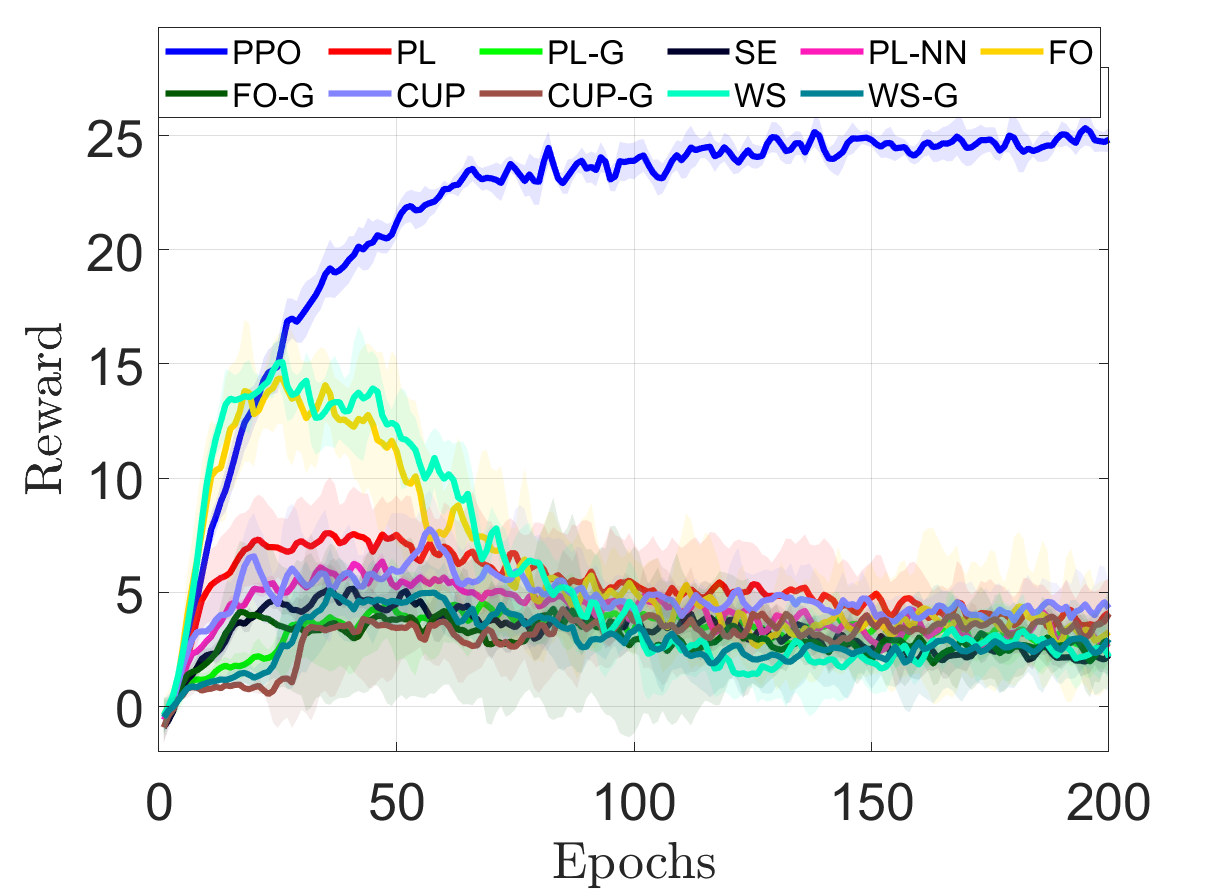}\\[-9pt]
\includegraphics[trim={7mm 0mm 7mm 3mm}, width=.22\linewidth]{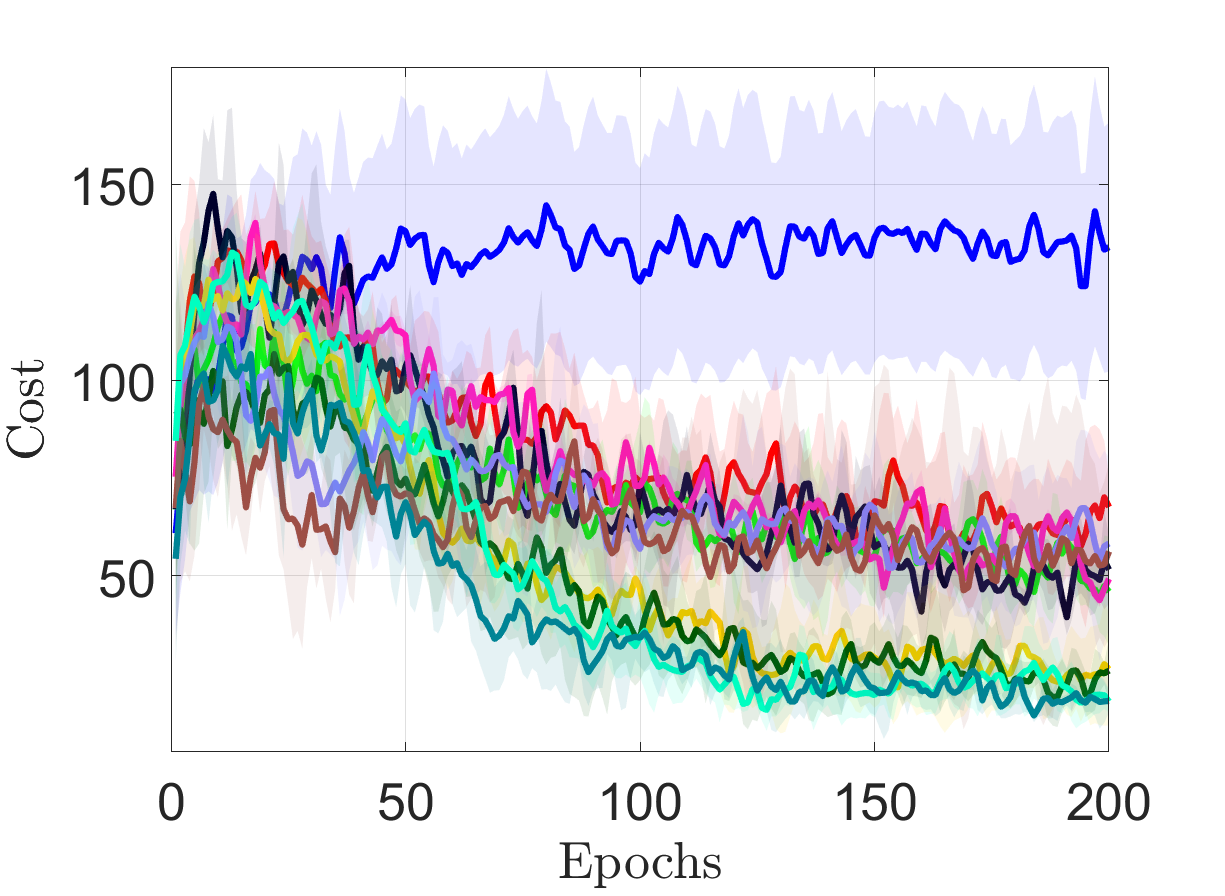}
\end{tabular}%
\label{fig.results_button2}}
\hfil
\subfloat[Goal]{\begin{tabular}[b]{c}
\includegraphics[trim={7mm 0mm 7mm 3mm}, width=.22\linewidth]{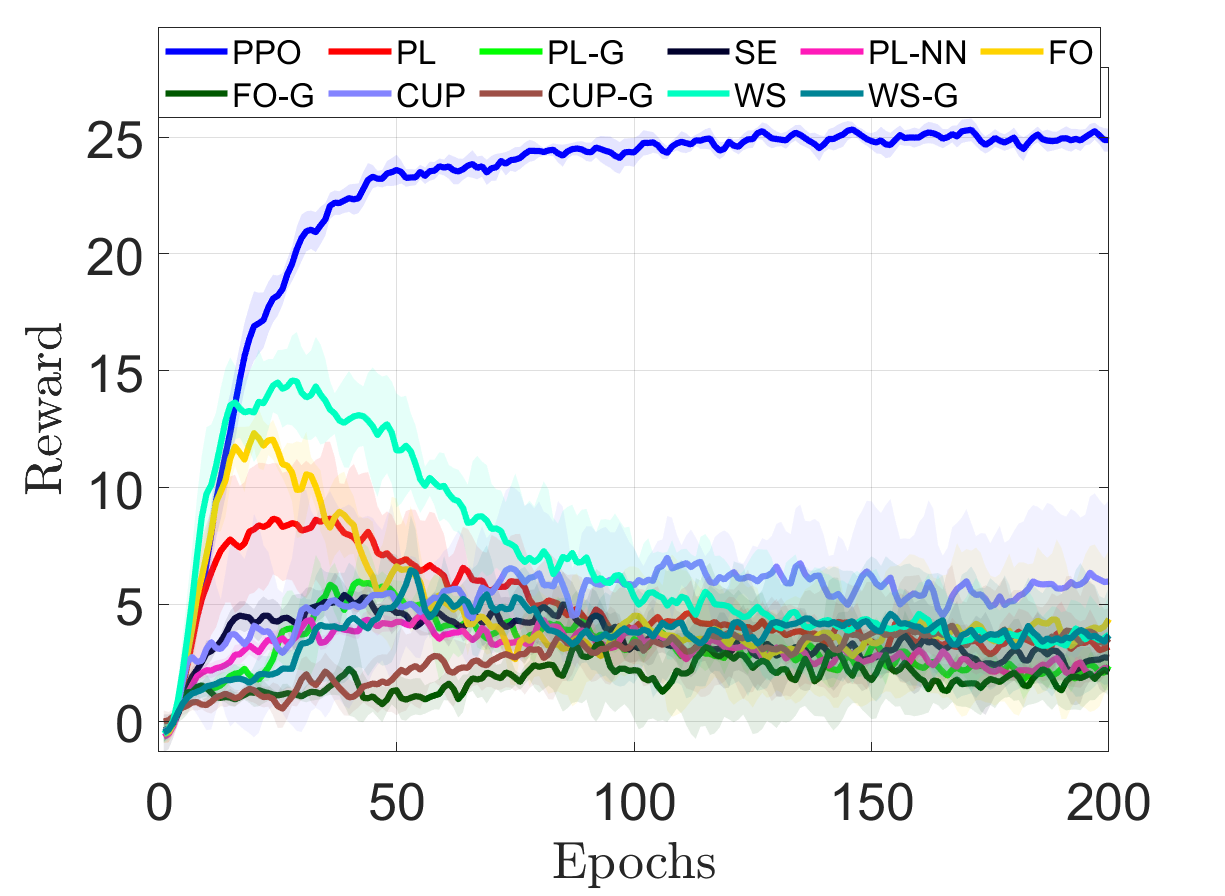}\\[-9pt]
\includegraphics[trim={7mm 0mm 7mm 3mm}, width=.22\linewidth]{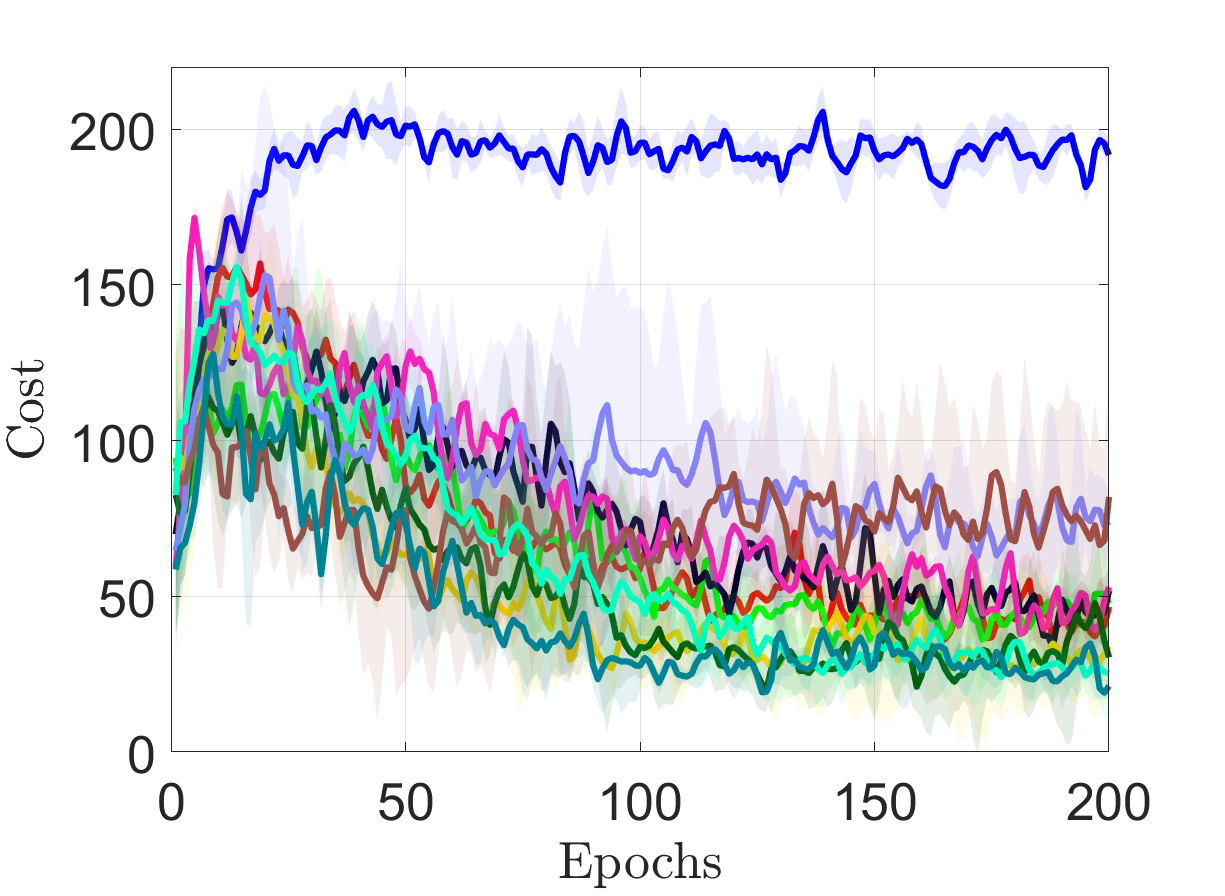}
\end{tabular}%
\label{fig.results_goal2}}
\hfil
\subfloat[Push]{\begin{tabular}[b]{c}
\includegraphics[trim={7mm 0mm 7mm 3mm}, width=.22\linewidth]{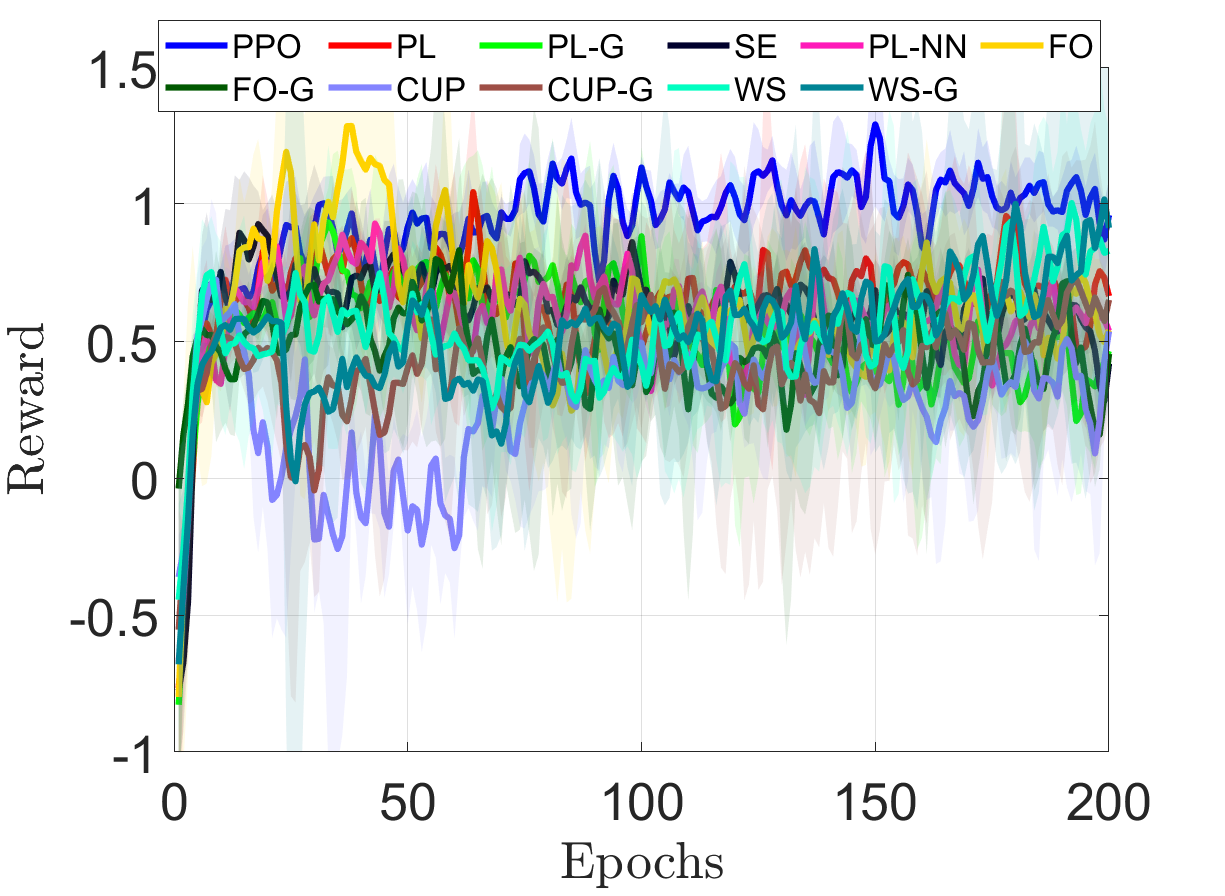}\\[-9pt]
\includegraphics[trim={7mm 0mm 7mm 3mm}, width=.22\linewidth]{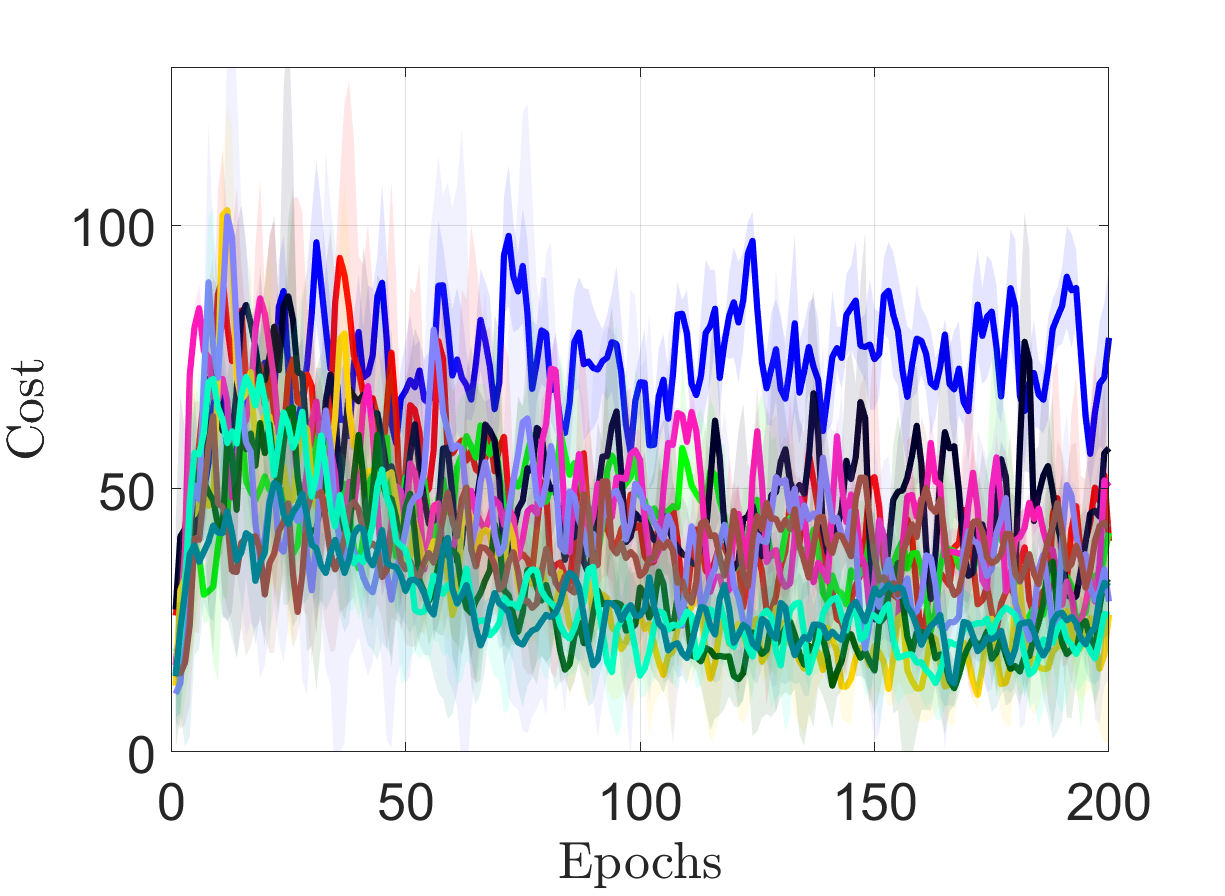}
\end{tabular}%
\label{fig.results_push2}}
\hfil
\subfloat[Swimmer]{\begin{tabular}[b]{c}
\includegraphics[trim={7mm 0mm 7mm 3mm}, width=.22\linewidth]{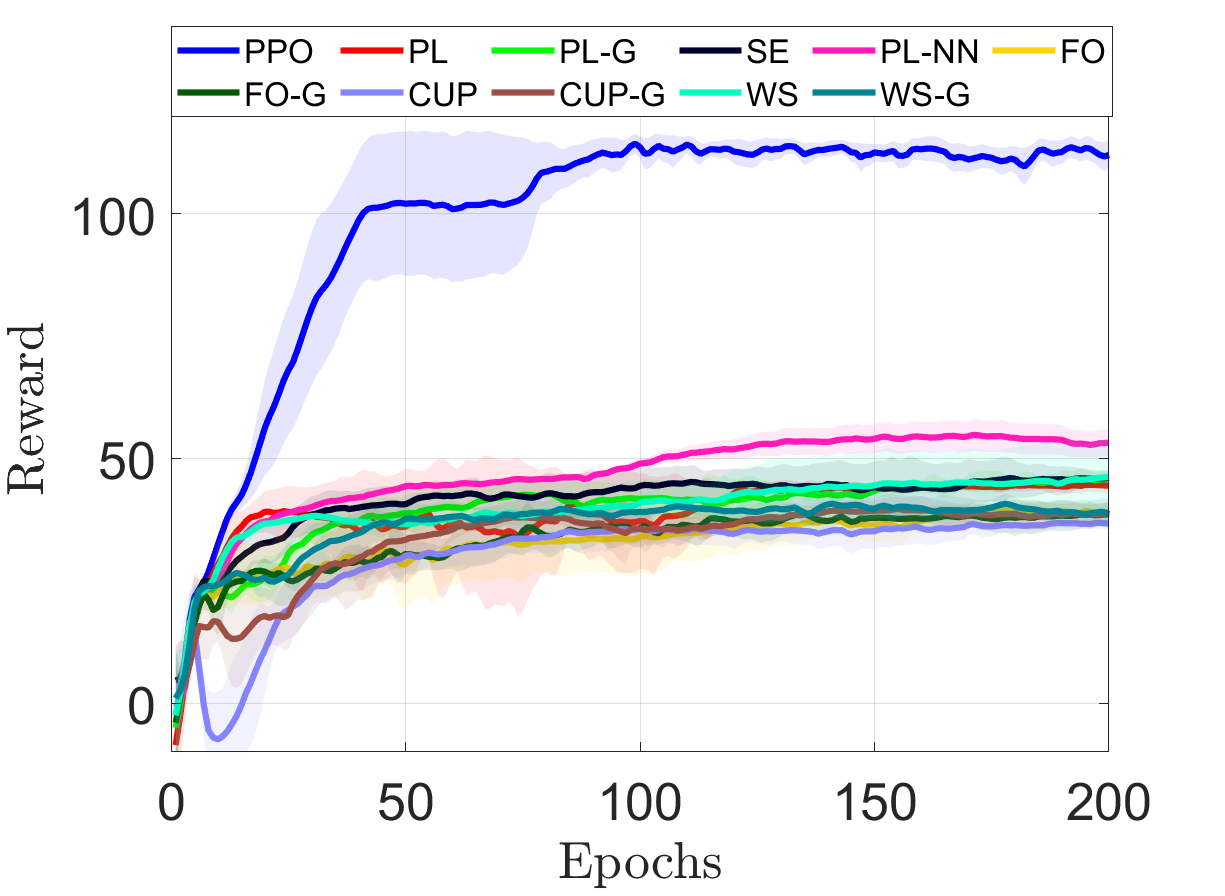}\\[-9pt]
\includegraphics[trim={7mm 0mm 7mm 3mm}, width=.22\linewidth]{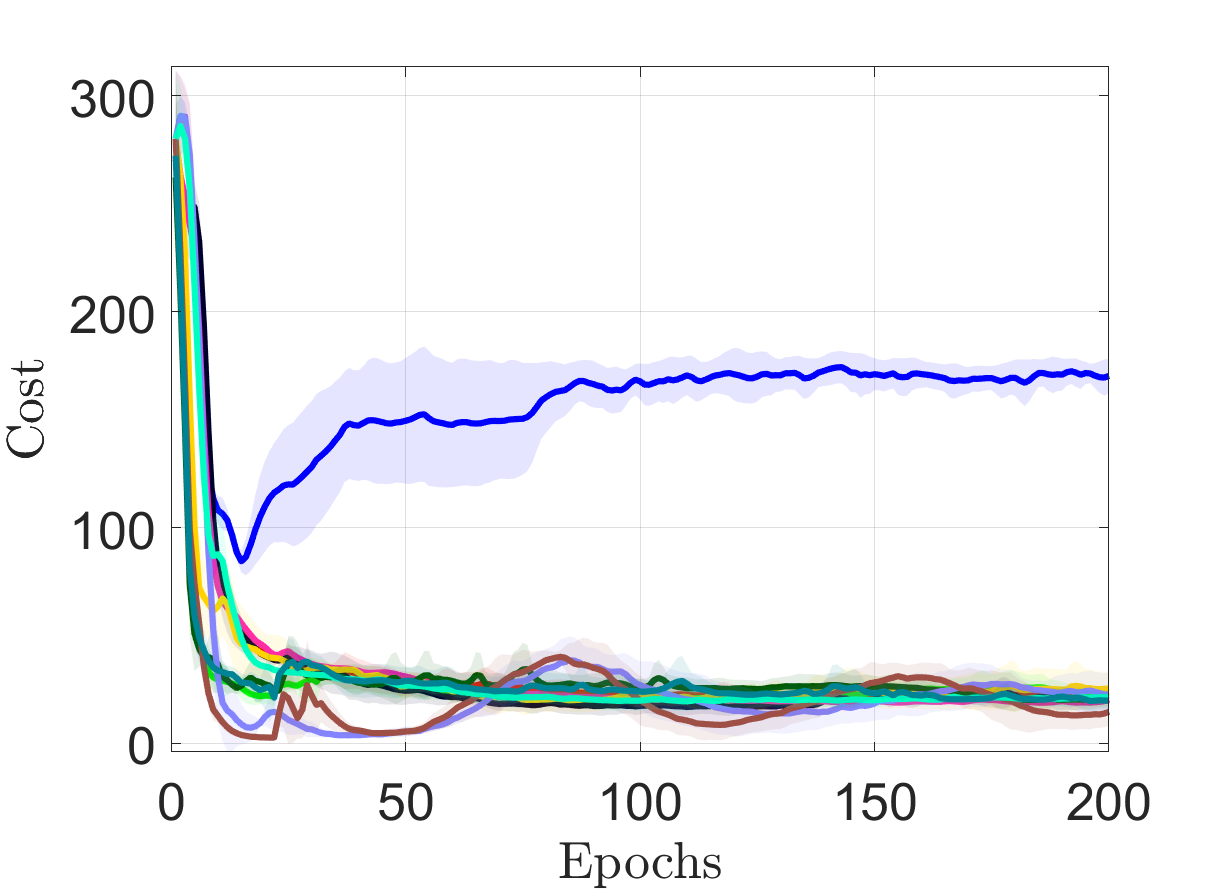}
\end{tabular}%
\label{fig.results_swimmer}}
\hfil
\subfloat[Hopper]{\begin{tabular}[b]{c}
\includegraphics[trim={7mm 0mm 7mm 3mm}, width=.22\linewidth]{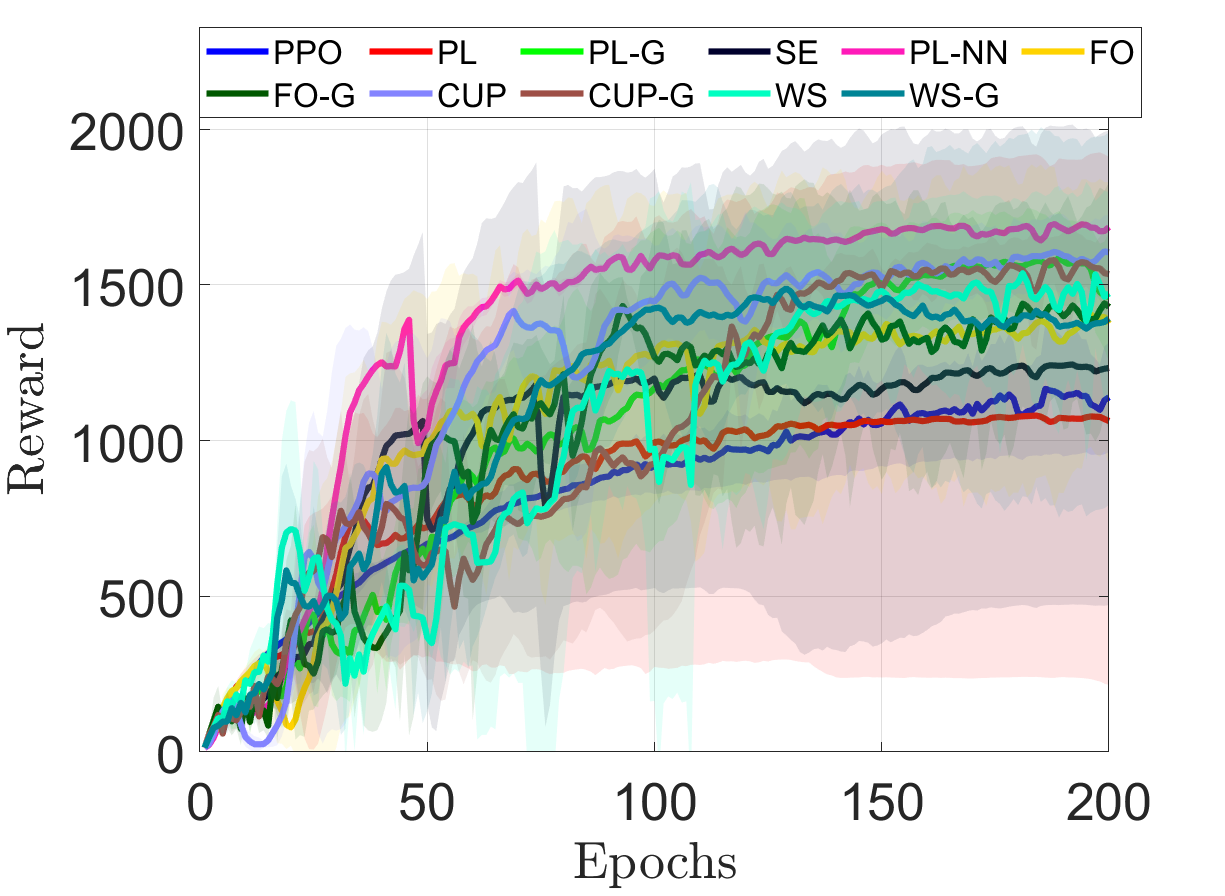}\\[-9pt]
\includegraphics[trim={7mm 0mm 7mm 3mm}, width=.22\linewidth]{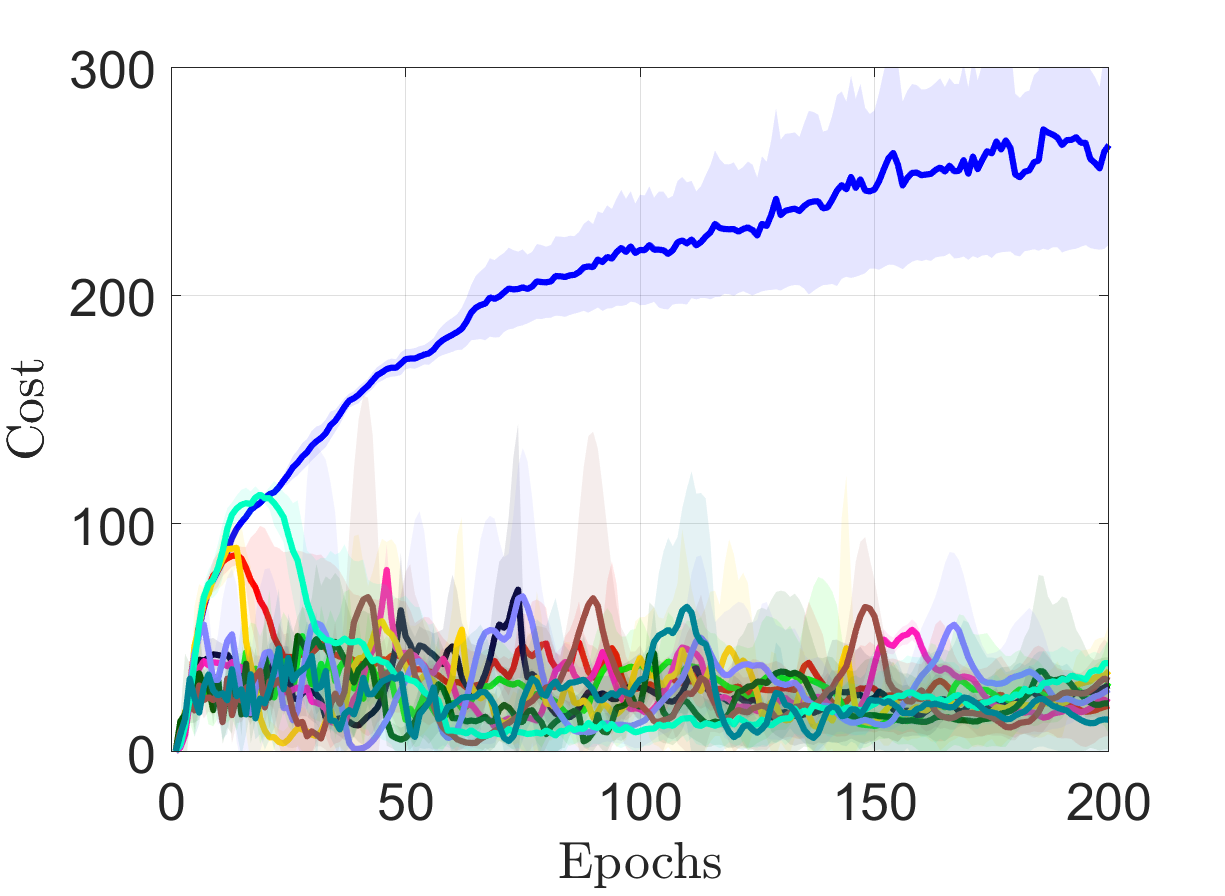}
\end{tabular}%
\label{fig.results_hopper}}
\caption{Reward and cost returns throughout the learning process for different learning methods across the eight SRL tasks. A more detailed version of the reward and cost plots is provided in the supplementary material.}
\label{fig.rewards_costs}
\vspace{-10pt}
\end{figure*}

We also compare the performance of different learning methods by evaluating the reward and cost returns throughout the learning process. 
The reward indicates how effectively the trained control policy accomplishes the desired task, with a higher reward representing better task performance. 
Conversely, the cost reflects whether the system encounters unsafe states, thereby measuring safety performance. 
Higher costs signify the presence of more dangerous behaviors.

The results are presented in Fig.~\ref{fig.rewards_costs}. 
Due to page limit, more detailed individual comparison plots are provided in the supplementary material.
It can be observed that, compared to the PPO algorithm, which does not incorporate explicit safety mechanisms, all other approaches achieve a substantial reduction in cost returns throughout the learning phase.
Moreover, the proposed GenSafe greatly enhances the safety performance of existing SRL methods (see PL-G vs. PL, FO-G vs. FO, CUP-G vs. CUP, and WS-G vs. WS), achieving notable cost reductions during the early learning stages. 
This aligns with GenSafe's design objectives.
However, the effectiveness of GenSafe is influenced by the task complexity. 
For more challenging tasks, e.g., the \textit{Push} task, ensuring a safer learning process becomes more difficult, resulting in lesser cost reductions.
Nonetheless, the algorithms augmented by GenSafe consistently outperform the original SRL algorithms in terms of safety performance.
Additionally, due to the aforementioned challenges in generating reliable cost estimates, both the SE and PL-NN methods exhibit diminished performance compared to the PL-G algorithm.
This further validates the usefulness and effectiveness of our proposed GenSafe.

Meanwhile, the algorithms augmented by GenSafe achieve similar final rewards to their respective original versions, albeit at a slower pace. 
A primary factor contributing to the slower increase in reward is that, to realize a safer learning process, GenSafe restricts the exploration of learning algorithms to regions deemed reliably safe.
While this improves safety performance, it naturally reduces the speed of reward growth, as the search areas for potentially higher-reward policies are constrained.
This phenomenon is also observable when comparing PPO to other SRL algorithms, which, despite achieving higher reward returns, does so at the expense of safety due to its unrestricted exploration.
Such a trade-off between safety and task performance is recognized in various SRL studies~\cite{zhou2020general,brunke2022safe,gu2022review}.
Nevertheless, GenSafe is still able to achieve final rewards comparable to those of the original SRL algorithms, demonstrating its ability to enhance safety while maintaining task performance at a satisfactory level.

\section{Discussion}
\label{sec.discussion}

In this section, we discuss several important aspects of the proposed approach, as well as its limitations and possible future directions.



\subsection{Trade-off between Safety and Task Performance}

As aforementioned, determining the trade-off between maximizing reward returns and ensuring system safety is a challenging problem in SRL.
In the proposed GenSafe, this trade-off is modifiable through parameters such as the estimated cost for unobserved reduced state-action pairs $\Delta$, the immediate cost threshold $d_s$, and the future cost threshold $d$.
Setting a lower estimated cost $\Delta$ tends to classify more unexplored regions as safe.
Then, together with higher thresholds for $d_s$ and $d$, a more expansive exploration is allowed, thereby leading to a more efficient policy search and improved rewards.
However, this increased exploratory flexibility also raises the possibility of encountering unsafe behaviors, which, in turn, results in higher cost returns.
How to determine the optimal balance between safety and task performance remains an open research question.
Currently, decisions often rely on experience-driven manual adjustments that are based on prior knowledge of system characteristics and task requirements. 
Further research is needed to develop more reliable and systematic approaches for managing this trade-off.

\subsection{Limitations}

One major limitation of the proposed GenSafe is the increased computational demand stemming from activities like constructing the ROMDP and performing action corrections. 
Although parallel computing techniques can be used to mitigate this issue, they inevitably prolong the training duration and increase the requirements for computational resources.
One possible solution to this problem is to reformulate the constraints within the optimization~\eqref{eq.action_correction} into differentiable forms, thereby enabling the usage of gradient-based methods or the derivation of analytical solutions.
However, devising a method for accurately and reliably transforming these constraints needs more investigation.
Another limitation of GenSafe, echoing a common issue among model-free SRL algorithms, is the challenge of providing absolute safety guarantees, i.e., complete avoidance of unsafe behaviors.
Given its reliance on a data-driven approach, the algorithm has to encounter unsafe regions to identify and subsequently avoid them, which inevitably leads to unsafe actions during the learning phase.
To achieve absolute safety guarantees, model-based SRL methods that leverage precise system models and control-theoretical concepts may offer a viable path. 
Nevertheless, this area remains open for in-depth research and development.

\subsection{Future Work}
For future work, we plan to extend the current approach to policy-level modifications. 
This could be done by, e.g., integrating the cost predictions generated by the ROMDP into the policy update mechanism of the selected SRL algorithm.
However, this necessitates improving the ROMDP's long-term prediction accuracy to ensure a reliable policy-level adjustment.
Another direction involves exploring the use of different order reduction techniques in constructing the ROMDP, e.g.~\cite{zhang2022towards,yang2021causalvae,uehara2021representation}.
In this work, we employ t-SNE primarily due to the reason that it is a purely data-driven approach, thereby imposing no prerequisites on the properties of the system or control policy. 
This greatly enhances the applicability of the proposed approach, making it suitable for a wide range of SRL tasks and algorithms. 
However, a potential drawback of t-SNE is that it might lead to substantial loss of information during the reduction process, which may limit the performance of GenSafe. 
Utilizing a more efficient and reliable order reduction technique could potentially improve the overall performance of our proposed approach.

\section{Conclusion}
\label{sec.conclusion}

In this work, we present the Generalizable Safety Enhancer, referred to as GenSafe, for improving the safety performance of SRL algorithms.
Leveraging data collected throughout the learning process, we first apply a model order reduction technique to transform the original high-dimensional state space into a representative two-dimensional state space.
Then, based on this reduced state space, we construct a ROMDP through a comprehensive abstraction process that includes state, action, cost, transition, and policy abstractions. 
The resultant ROMDP serves as a low-dimensional approximator of the original cost function in CMDP.
Subsequently, by reformulating the original cost constraint into ROMDP-based constraints, we propose an action correction mechanism that adjusts each applied action to enhance the probability of constraint satisfaction.
Through this action-level correction strategy, the proposed GenSafe acts as an additional safety layer for SRL algorithms, providing a broad compatibility with a wide range of SRL algorithms.
We evaluate the performance of GenSafe across various SRL benchmark problems.
The results illustrate the effectiveness of GenSafe in improving the safety performance of SRL algorithms, especially in early learning stages, while also maintaining the task performance at a satisfactory level.
This validates the utility and applicability of GenSafe, suggesting its potential as a versatile tool for enhancing safety across diverse SRL applications and scenarios.
For future work, we aim to explore extensions involving policy-level modifications and alternative order reduction techniques, thereby further enhancing the performance and generalizability of GenSafe.




\bibliographystyle{IEEEtran}
\bibliography{ref.bib}

 




\begin{IEEEbiography}[{\includegraphics[width=1in,height=1.25in,clip,keepaspectratio]{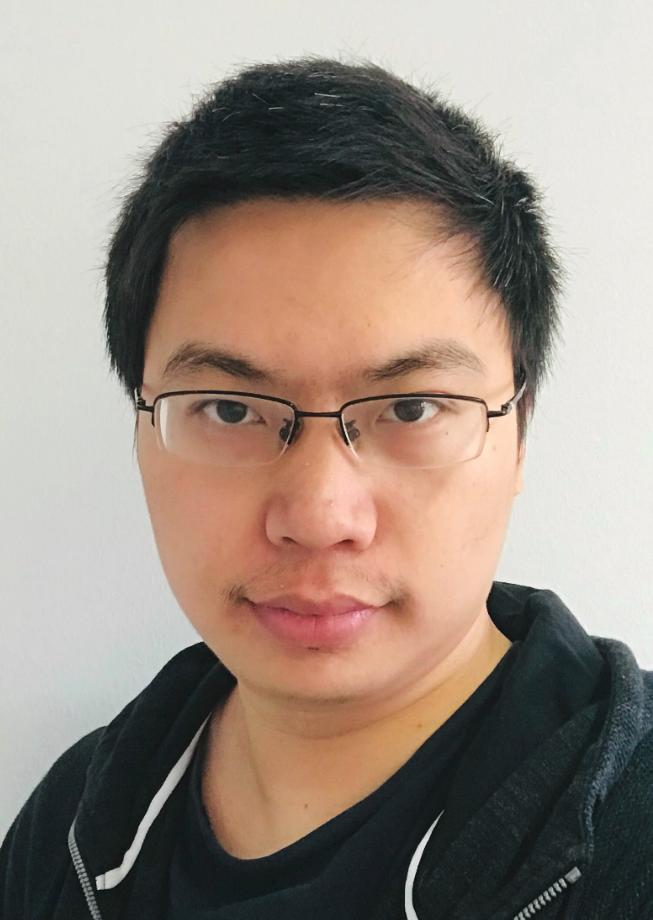}}]{Zhehua Zhou} 
received the B.E. degree in mechatronics engineering from Tongji University, Shanghai, China, in 2014, and the M.Sc. degree in electrical and computer engineering from the Technical University of Munich, Munich, Germany, in 2017. In 2022, he received the Ph.D. degree in electrical and computer engineering from the Technical University of Munich, Munich, Germany. He is currently working as a Postdoctoral Research Fellow at Department of Electrical and Computer Engineering, University of Alberta, Edmonton, Canada. His research interests include optimal control, learning-based control, and applications to robotics.
\end{IEEEbiography}

\begin{IEEEbiography}[{\includegraphics[width=1in,height=1.25in,clip,keepaspectratio]{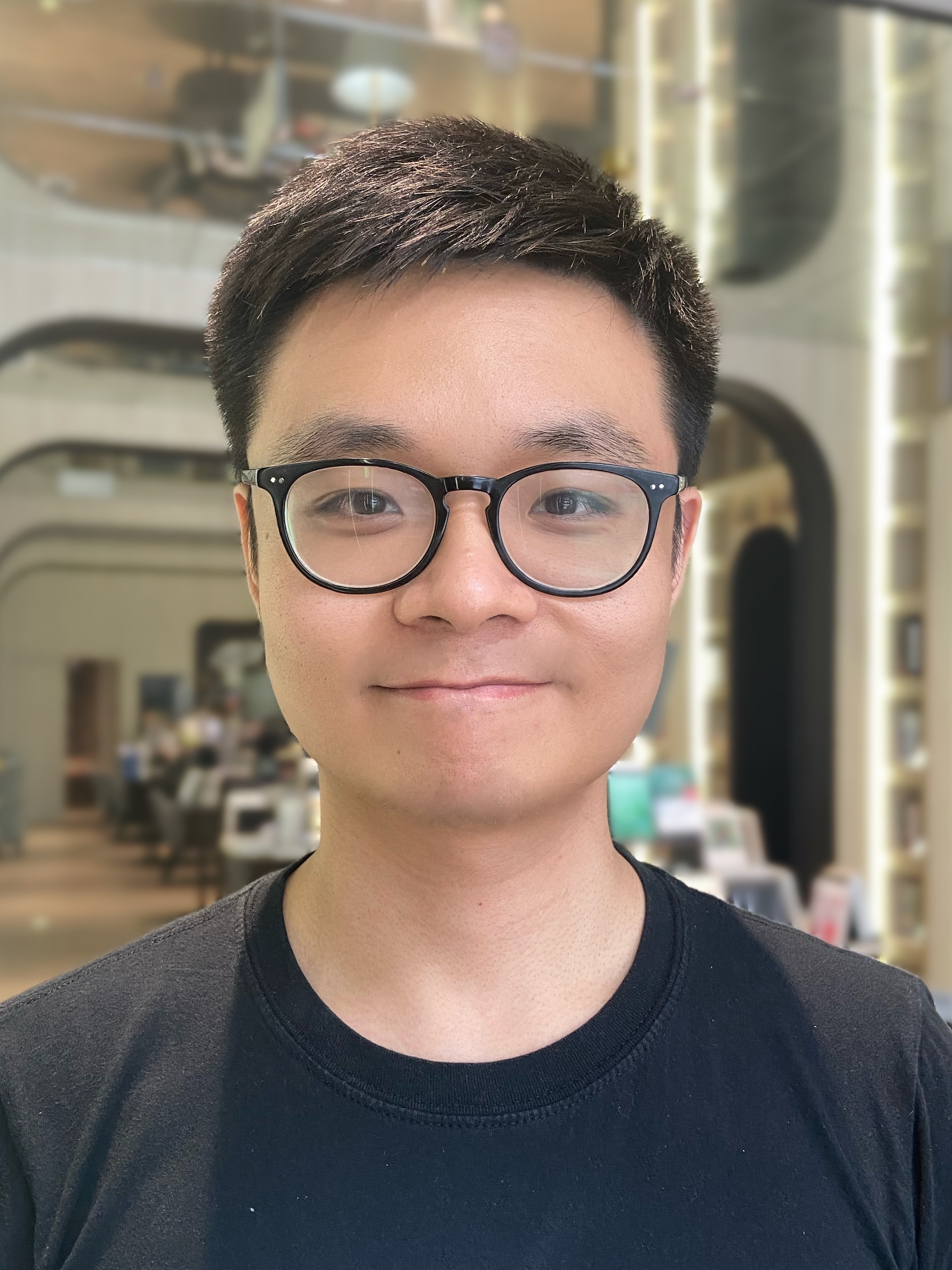}}]{Xuan Xie} 
is a Ph.D. student at the University of Alberta, Canada. He received the B.E. degree from Sun Yat-sen University, China, and the master’s degree from Max Planck Institute for Software Systems, Germany. His research interests include trustworthiness analysis for large language model, safety analysis of AI-enabled cyber-physical systems, and neural network verification.
\end{IEEEbiography}

\begin{IEEEbiography}[{\includegraphics[width=1in,height=1.25in,clip,keepaspectratio]{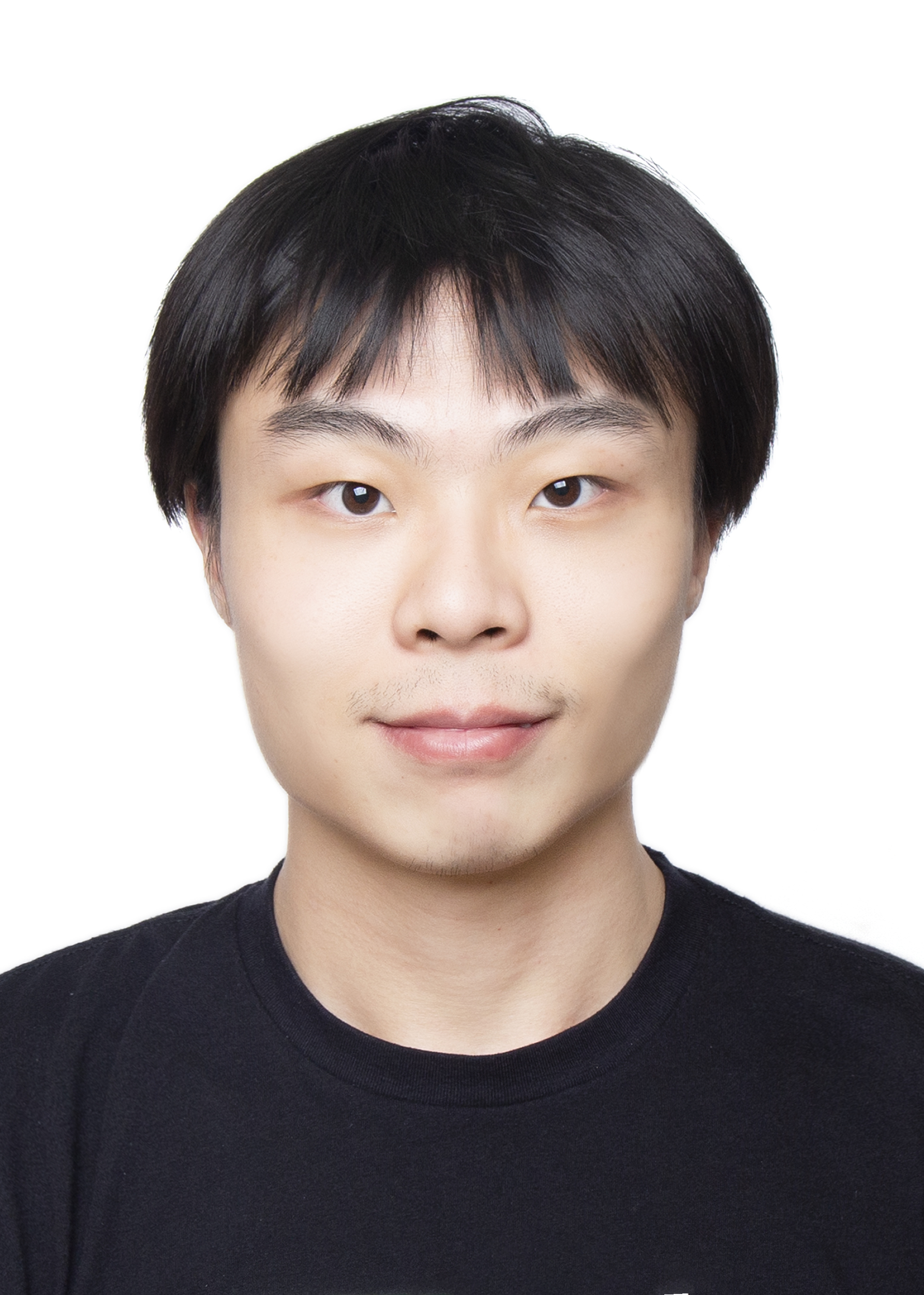}}]{Jiayang Song} 
is a PhD candidate at University of Alberta, Canada. He received his B.E. from Western University, Canada and his M.E. from University of Toronto, Canada. His research topics include testing, analysis, repairing and enhancement of AI systems and their applications for quality assurance of trustworthy AI-enabled cyber-physical systems.
\end{IEEEbiography}

\begin{IEEEbiography}[{\includegraphics[width=1in,height=1.25in,clip,keepaspectratio]{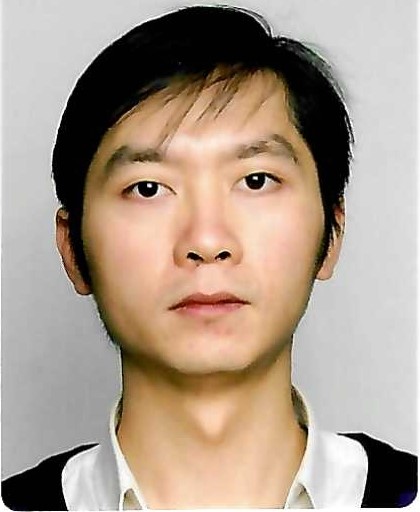}}]{Zhan Shu} 
received his B.Eng. degree in Automation from Huazhong University of Science and Technology in 2003, and his PhD degree in Control Engineering from The University of Hong Kong in 2008. 
He was a Postdoctoral Fellow at the Hamilton Institute, the National University of Ireland, Maynooth, from 2009 to 2011 and a Lecturer in the Faculty of Engineering and Physical Sciences, the University of Southampton from 2011 to 2019. Now, he is a Professor in the Department of Electrical and Computer Engineering, the University of Alberta. He is a Senior Member of IEEE, a Member of IET, and an invited reviewer of the Mathematical Review of the American Mathematical Society. He serves as an Associate Editor for Mathematical Problems in Engineering, Asian Journal of Control, Journal of The Franklin Institute, Proc. IMechE, Part I: Journal of Systems and Control Engineering, IET Electronics Letters, IET Control Theory and Applications, IEEE Trans. Automatic Control, and a member of the IEEE Control Systems Society Conference Editorial Board. His current research interests include networked control, learning-based control, distributed systems, and hybrid systems.
\end{IEEEbiography}

\begin{IEEEbiography}[{\includegraphics[width=1in,height=1.25in,clip,keepaspectratio]{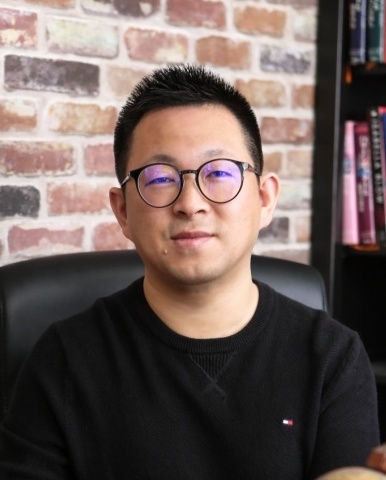}}]{Lei Ma} 
is currently an associate professor at The University of Tokyo, Japan and University of Alberta, Canada. He was honorably selected as a Canada CIFAR AI Chair and Fellow at Alberta Machine Intelligence Institute (Amii). Previously, he received the B.E. degree from Shanghai Jiao Tong University, China, and the M.E. and Ph.D. degrees from The University of Tokyo, Japan. His recent research centers around the interdisciplinary fields of software engineering (SE) and trustworthy artificial intelligence, with a special focus on the quality, reliability, safety and security aspects of AI Systems.
\end{IEEEbiography}

\vfill

\end{document}